\documentclass[sigconf,authorversion,nonacm]{acmart}

\usepackage{graphicx}
\usepackage{booktabs} %

\usepackage[utf8]{inputenc} %
\usepackage[T1]{fontenc}    %
\usepackage{xcolor}         %
\usepackage{amsmath}
\usepackage[frozencache=true,cachedir=.]{minted}
\usepackage{wrapfig}
\usepackage{todonotes}
\usepackage{placeins}
\usepackage[most]{tcolorbox}
\usepackage{caption}
\usepackage{subcaption}
\usepackage{fontawesome5}
\usepackage{multirow}

\usepackage{hyperref}
\AtBeginDocument{%
  \providecommand\BibTeX{{%
    \normalfont B\kern-0.5em{\scshape i\kern-0.25em b}\kern-0.8em\TeX}}}

\acmConference{}{}{}

\begin{document}

\title{A Data-Centric Optimization Framework for Machine Learning}

\author{Oliver Rausch}
\affiliation{%
  \institution{Department of Computer Science}
  \city{ETH Zurich}
  \country{Switzerland}
}
\email{oliver.rausch@inf.ethz.ch}
\authornote{Both authors contributed equally to the paper}

\author{Tal Ben-Nun}
\authornotemark[1]
\affiliation{%
  \institution{Department of Computer Science}
  \city{ETH Zurich}
  \country{Switzerland}
}
\email{talbn@inf.ethz.ch}

\author{Nikoli Dryden}
\affiliation{%
  \institution{Department of Computer Science}
  \city{ETH Zurich}
  \country{Switzerland}
}
\email{nikoli.dryden@inf.ethz.ch}

\author{Andrei Ivanov}
\affiliation{%
  \institution{Department of Computer Science}
  \city{ETH Zurich}
  \country{Switzerland}
}
\email{anivanov@inf.ethz.ch}

\author{Shigang Li}
\affiliation{%
  \institution{Department of Computer Science}
  \city{ETH Zurich}
  \country{Switzerland}
}
\email{shigang.li@inf.ethz.ch}

\author{Torsten Hoefler}
\affiliation{%
  \institution{Department of Computer Science}
  \city{ETH Zurich}
  \country{Switzerland}
}
\email{htor@inf.ethz.ch}

\renewcommand{\shortauthors}{Rausch et al.}

\begin{abstract}
Rapid progress in deep learning is leading to a diverse set of quickly changing models, with a dramatically growing demand for compute.
However, as frameworks specialize performance optimization to patterns in popular networks, they implicitly constrain novel and diverse models that drive progress in research. 
We empower deep learning researchers by defining a flexible and user-customizable pipeline for optimizing training of arbitrary deep neural networks, based on data movement minimization. 
The pipeline begins with standard networks in PyTorch or ONNX and transforms computation through progressive lowering. We define four levels of general-purpose transformations, from local intra-operator optimizations to global data movement reduction. These operate on a data-centric graph intermediate representation that expresses computation and data movement at all levels of abstraction, including expanding basic operators such as convolutions to their underlying computations.
Central to the design is the interactive and introspectable nature of the pipeline. Every part is extensible through a Python API, and can be tuned interactively using a GUI. We demonstrate competitive performance or speedups on ten different networks, with interactive optimizations discovering new opportunities in EfficientNet.

\end{abstract}

\maketitle

\section{Introduction}
\label{sec:intro}
The modern development of deep learning is spearheaded by the conflux of algorithms~\cite{bottou2018optimization},
data~\cite{halevy2009unreasonable,sun2017revisiting}, hardware~\cite{raina2009large,jouppi2017datacenter}, and systems~\cite{tensorflow2015-whitepaper,paszke2019pytorch}.
Today, machine learning systems have become critical in light of the increasing complexity of models and the breadth of emerging hardware.
Training performance in particular is key to both researcher productivity and reducing the environmental impact of machine learning~\cite{strubell2019energy,patterson2021carbon}.
To this end, systems and compilers such as XLA~\cite{xla}, ONNX Runtime~\cite{onnxruntime}, and TorchScript~\cite{torchscript} are widely used to automatically optimize training.

Most such compilers tend to focus on \textit{operator-centric} optimization, that is, defining specific rule-sets based on a set of predefined building blocks.
Thus, these frameworks tend to perform very well on popular neural networks, but often have limited performance improvements on new, unseen models --- which many researchers would like to explore, but may not be able to train given the high compute costs.
Indeed, this effect has led to a situation where the models that perform well on existing hardware and systems are more likely to succeed than others~\cite{hooker2020hardware}.

In this work, we propose to shift the paradigm from operators to their memory access patterns by performing data-centric Deep Neural Network (DNN) optimization. It is well-established that the performance of modern DNNs such as Transformers hinges on data movement minimization~\cite{ivanov2021data,deepspeed}, and that a certain FLOP reduction does not guarantee a matching speedup~\cite{tan2019efficientnet}. Up until now such optimizations have been performed by specialized engineering teams for specific architectures, remaining out of reach for most research groups~\cite{barham2019machine}.

We present \textbf{DaCeML}\footnote{\url{https://github.com/spcl/daceml}},
a Data-Centric Machine Learning framework, which provides a simple, flexible, and customizable Python-based pipeline for optimizing training and empowering deep learning research. DaCeML seamlessly integrates with PyTorch~\cite{paszke2019pytorch} and ONNX~\cite{onnx} to enable accelerating and tuning existing models, both in evaluation and backpropagation. The input models are then optimized with a general-purpose transformation pipeline, which reduces data movement at fine and coarse grain, regardless of the internal computation. Lastly, DaCeML provides interfaces to programmatically and interactively guide the optimization process further.

Internally, the framework uses Stateful Dataflow Multigraphs (SDFGs)~\cite{ben2019stateful} as a data-centric, hierarchical, graph-based intermediate representation, which enables it to work with operators and data movement at all levels, from registers to distributed memory.
The DaCeML optimization pipeline begins with a standard PyTorch or ONNX model and transforms computation through progressive lowering, using four levels of general-purpose transformations: coarse-grained; local data movement reduction; global data layout optimization; and hardware specialization. 
Data-centric optimizations manifest in different ways, especially when training is considered. One of the several unique controls DaCeML provides, for example, is choosing whether to recompute or retain data for backpropagation.
With the included automatic optimizations, DaCeML often \textit{already matches or outperforms modern frameworks}.

The central focus of DaCeML is the ability to then go further --- it unpacks the DNN compiler-black box and puts the machine learning practitioner in the driver's seat. This starts with the visual, introspectable intermediate representation that makes finding and understanding performance issues, such as excessive data movement, intuitive. These are subsequently addressed by manipulating data movement or the data layouts of the parameters and intermediate storage. The SDFG IR allows automated or manual searches to be performed on multiple granularities \textit{simultaneously}, rather than in the traditional compiler-based fixed set of passes.
These can be performed either using a Python API or interactively using the Visual Studio Code~\cite{vscode} IDE.

\begin{figure*}[t]
    \centering
    \includegraphics[width=0.9\linewidth,page=1]{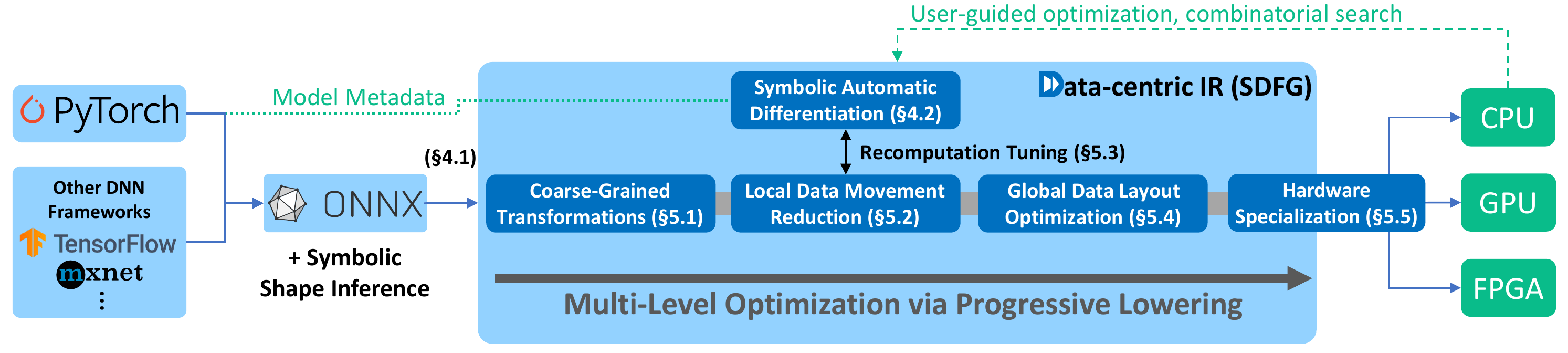}
    \vspace{-1.2em}
    \caption{DaCeML system overview.}\vspace{-0.75em}
    \label{fig:overview}
\end{figure*}

We provide an overview of DaCeML's design, its approach to progressive lowering, and its transformations, including novel optimizations not supported by other frameworks. In particular, we demonstrate up to 3.43$\times$ speedups over prior best on non mainstream activations; state-of-the-art performance with automatic optimization of a wide range of DNNs from various domains~\cite{yolov4,devlin2019bert,he2015deep,long2015fcn,naumov2019dlrm,oord2016wavenet,sandler2019mobilenetv2,tan2019efficientnet,tolstikhin2021mlpmixer,zagoruyko2017wide}, compared with PyTorch, TensorFlow + XLA, and JAX; and two guided optimization case studies on BERT~\cite{devlin2019bert} %
and EfficientNet~\cite{tan2019efficientnet}, the former highlighting the importance of data layout and the latter showing that nonlocal data movement minimization can yield 1.33$\times$ speedup over cuDNN.

We make the following contributions:
\begin{enumerate}
    \item Design and implementation of a framework for data-centric lowering and multi-level optimization of DNNs for training.
    \item A formulation of symbolic automatic differentiation (AD) on a data-centric intermediate representation, which enables novel
    automatic synthesis of backward-mode operators for arbitrary operators, including lowered kernels. 
    \item Integration of differentiation into the optimization pipeline, enabling joint optimization pre- and post-AD for data movement reduction and adaptive recomputation.
    \item Demonstration of a human-guided workflow for DNN compilation, in which performance engineers interactively guide compilation using a graphical interface. This achieves state-of-the-art performance on highly contested deep neural networks~\cite{devlin2019bert, tan2019efficientnet} and exposes previously untapped optimization opportunities.
\end{enumerate}

\section{Background and Related Work}
\label{subsec:related}
The data-centric view is garnering attention in the DNN optimization community.  \citet{ivanov2021data} consider a data movement minimization approach to optimize training of transformers~\cite{vaswani2017attention,devlin2019bert}, observing that the models are largely memory bound.
While their work, which focuses exclusively on BERT~\cite{devlin2019bert}, is entirely manual, it paves the way towards automating such analyses and optimization.
Similarly, DeepSpeed~\cite{deepspeed} also provides manually-optimized primitives for transformers.
Numerous other works have focused on specific optimizations~\cite{frostig2018compiling,jia2019optimizing,sivathanu2019astra,baghdadi2019tiramisu,vasilache2018tensor,lethin2019polyhedral,wei2017dlvm,truong2016latte,venkat2019swirl,dong2019acorns,elango2018diesel,hu2020featgraph,oyama2018accelerating,li2016optimizing,zheng2020ansor,li2020partialcoll,steiner2020value,yang2021equality} that can be implemented as data-centric transformations.

In the rest of this section, we describe the state of the practice in DNN optimization, and highlight important distinctions between current approaches and the DaCeML multi-level optimization pipeline for training.

\textbf{Training vs. inference compilation}\; Inference optimization elides several concepts required in training. Most importantly, no automatic differentiation is needed. Secondly, as opposed to training, intermediate activations need not be stored for backpropagation, which drastically changes the optimization search space. Thirdly, several operators (e.g., batch normalization, dropout, convolution) behave differently during training. For example, batch normalization stores running statistics throughout the process, and convolutions that use basis transformations (FFT, Winograd) can pre-transform the weights once, since they will not change during inference. Lastly, apart from backpropagation, gradient updates and stochastic optimizer rules must be applied.

\textbf{State-of-the-art DNN compilers}\; Researchers have often used custom kernels, linear algebra primitives (e.g., BLAS), and vendor-optimized libraries (e.g., cuDNN~\cite{chetlur2014cudnn}, oneDNN~\cite{onednn}).
Recently, with the modularization and diversity of DNNs, the focus is shifting towards using general compiler infrastructure in DNN optimization, which uses Just-in-Time (JIT) compilation to both optimize individual operators and fuse them.

The methods by which DNN compilers optimize (transform) code can be classified into three categories: \textit{graph rewriting rules}, \textit{expansions}, and \textit{global passes}.
Graph rewriting represents DNNs as DAGs and pattern-matches certain subgraphs to replace them with others.
Expansions convert a known operation (e.g., batch normalization) directly into an explicit, pre-optimized version.
Finally, global passes operate on the entire code (e.g., memory scheduling). Ultimately, the first two categories, and sometimes the third, are implemented on an operator-centric basis. Below, we discuss state-of-the-art DNN compilers and their inner workings.

\textbf{XLA}\; Advanced compiler infrastructure~\cite{xla} used by TensorFlow~\cite{tensorflow2015-whitepaper} and JAX~\cite{jax}, which contains multiple intermediate representations (e.g., HLO, LLO) and various domain-specific expansion transformations and global passes. Apart from those, general purpose optimizations (such as dead-code elimination) are performed by the LLVM~\cite{lattner2004llvm} and MLIR~\cite{lattner2020mlir} (experimental) infrastructure. Prior to XLA, graph rewriting rules in TensorFlow were provided by a component called Grappler.

\textbf{TorchScript} (\texttt{torch.jit})\; The main static optimization effort in PyTorch~\cite{rotem2018glow,torchscript,paszke2019pytorch}. Through either tracing or Python introspection, TorchScript can convert PyTorch modules into a DAG-based IR, and contains a pass manager to manage transformations. As opposed to XLA, optimizations are focused on the forward pass, leaving backpropagation to the \texttt{autograd} module. %

\textbf{ONNX Runtime}\; ONNX Runtime~\cite{onnxruntime} provides graph rewrite rules and  global passes, mostly focused on cleaning up artifacts resulting from the ONNX~\cite{onnx} format (e.g., constant folding) and algebraic fusion involving chained ONNX operators.%

\textbf{Inference frameworks}\; Frameworks that maintain a graph structure also provide optimizing transformations. 
TVM~\cite{chen2018tvm} provides passes on its two IRs (Relay, TIR) based on statement visitors/mutators, similar to AST manipulation.
TASO~\cite{jia2019taso} also provides subgraph pattern rewrite rules, focused on linear algebra.
OpenVINO~\cite{openvino} uses nGraph~\cite{cyphers2018intel} with graph rewrite rules and single-node matching transformations to optimize inference. For interactivity, it allows users to deselect certain optimization passes on given node names through its command-line optimizer.
TensorRT~\cite{tensorrt} and cuDNN~\cite{chetlur2014cudnn} provide functionality to fuse elementwise operations to DNN primitives, the former for the purpose of accelerating inference and the latter for training as well.

\textbf{Limitations}\; Each of the aforementioned DNN compilers is affected by at least one of three limiting factors.
First, most of the transformations rely on specific operator types, which change from network to network, and over time. 
Second, optimization of the backward pass and prior to differentiation are not well-explored in tandem, missing important potential reductions in data movement.
Lastly, DNN compiler optimizations are usually applied in a fixed set of passes, inhibiting opportunities for further tuning and analysis.

\section{System Overview}
\label{sec:system}
\begin{figure*}
    \centering
    \includegraphics[width=0.9\linewidth,page=2]{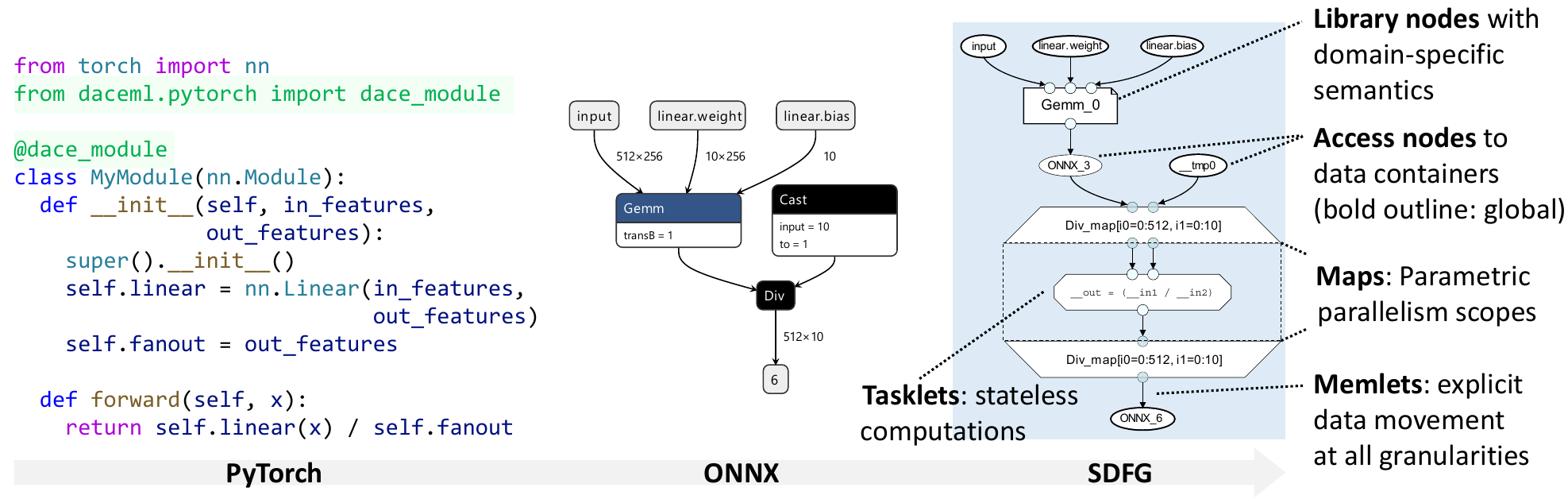}
    \vspace{-1em}
    \caption{SDFG Intermediate Representation and conversion. ONNX graph rendered in Netron~\cite{netron}.}
    \label{fig:sdfg}
\end{figure*}

We propose to tackle the problem of DNN optimization through a holistic and user-centered approach. The Data-Centric Machine Learning framework (\textbf{DaCeML}) is a semi-automated, user-extensible pipeline for compiling and optimizing deep learning workloads. It takes PyTorch models and ONNX files, optimizes the program through data-centric means, and generates code with state-of-the-art performance for multiple platforms.
The goal of DaCeML is threefold: (1)~\textbf{Usability}: keeping the simplicity and expressiveness of PyTorch and its powerful model definition; (2) \textbf{Generality:} using general-purpose transformations that generalize well to new models, from coarse-grained optimizations to low-level code generation; and (3) \textbf{Interactivity}: translating the simplicity of defining models to optimizing them by enabling easy extensibility and interactive reasoning.

To achieve this goal, DaCeML defines a \textit{progressive lowering} pipeline, depicted in Figure~\ref{fig:overview}. DaCeML uses the data-centric Stateful DataFlow multiGraph (SDFG) representation~\cite{ben2019stateful} --- a parametric graph IR designed for high-performance computing --- as its intermediate representation. SDFGs promote separation of concerns between domain scientists and performance engineers by optimizing the data movement of a program separately from the algorithmic part of the input code. The IR and associated transformations are written in Python and have  recently powered multiple applications in several domains, including quantum chemistry~\cite{ziogas2019optimizing} and numerical weather prediction~\cite{de2020stencilflow}, achieving state-of-the-art performance on CPUs, GPUs, and FPGAs.
This data-centric IR is fundamentally different from the existing task graph and multi-level representations. As opposed to LLVM (which uses opaque getelementptr accesses) and MLIR/XLA (which use memrefs that must be propagated globally for analysis), SDFGs enable \emph{constant-time evaluation} of all data movement in a graph (array access indices, sub-tensors, etc.) as symbolic expressions.

Once within that representation, DaCeML can automatically optimize the forward or backward passes of a model to performance that is on par with (or faster than) state-of-the-art DNN compilers.
Since most transformations operate on the simple elements of the SDFG IR (memlets, maps, tasklets, and access nodes) rather than coarse-grained operators, optimization generalizes across different operators. 

Furthermore, DaCeML's API is geared towards extensibility --- from defining new operator implementations, through local and global data movement planning, to low-level work partitioning in code generation --- allowing users to assume direct control and guide the optimization process towards faster performance. The process can be done interactively, through APIs and IDE support, or through combinatorial searches over the optimization space.%

\section{Data-Centric Progressive Lowering}
\label{sec:lowering}

DaCeML supports loading models from the Open Neural Network eXchange (ONNX) format~\cite{onnx}, or by wrapping PyTorch \texttt{nn.Module}s with a \texttt{@dace\_module} decorator. The loaded models can be invoked directly, or used as optimized replacements of sub-networks in existing training and inference codes. Figure~\ref{fig:sdfg} provides a summary of the conversion process and SDFG representation.

\subsection{From PyTorch/ONNX to Symbolic Dataflow}

The ONNX format is a DAG representation of DNNs, which aims to standardize deep learning primitives across frameworks and networks. In the representation (Figure~\ref{fig:sdfg}, middle), nodes represent operators and edges represent tensors that form data dependencies between them. As of version 1.9, there are 186 operator types in ONNX~\cite{onnx}, which cover the vast majority of deep learning models.

Most deep learning frameworks, including PyTorch, have converters to the ONNX representation (typically used for deployment). For this reason, we chose to build DaCeML around the ONNX representation. When a PyTorch module is wrapped, we use PyTorch's ONNX exporter to generate the computational graph. As this is not sufficient for training and parameter management, we augment the ONNX input with metadata such as positional arguments, module hierarchy, and parameter structure. From those elements, we generate a replacement \texttt{nn.Module} object that is controlled by DaCeML and seamlessly integrates with the PyTorch \texttt{autograd} module for training. Internally, the module keeps either one (inference) or two (forward, backward) SDFGs for execution and optimization.

\textbf{SDFG}\; We use the example in Figure~\ref{fig:sdfg} to briefly explain the SDFG representation. Full details and operational semantics for SDFGs are provided by \citet{ben2019stateful}.
In particular, SDFGs are graphs of graphs, representing state machines of acyclic dataflow multigraphs. Each \textit{state} in the state machine (light blue region) represents data access and computations (the equivalent of the full ONNX graph); the outer, state machine graph describes coarse-grained control flow (e.g., the training loop).

Two central concepts in SDFGs are the explicit separation of data movement from computation, and a multi-level view of data movement. The first concept is aided by the structure of the state graph itself --- computations (\textit{tasklets}, octagonal nodes) and \textit{access} (oval nodes) to data containers are represented by nodes, and data movement is represented explicitly by the edges, called \textit{memlets}. A tasklet cannot access non-local memory that is not connected to it via a memlet, and memlets are represented as symbolic ranges that connect access nodes and tasklets. Data container names are unique, and indicate disjoint memory regions, but access nodes that refer to them can appear multiple times. Furthermore, containers that are not local to the computation (i.e., cannot be removed in subsequent transformations) are called global memory and visually marked with a bold outline.

The concept of multi-level data movement is facilitated by three other types of nodes: (1) computations with predefined semantics are represented by \textit{library nodes} (e.g., matrix multiplication generated from the \texttt{Linear} layer in the figure); (2) parallel regions, such as the \texttt{Div} operator in the figure, are grouped together between \textit{map} (trapezoidal) nodes, which contain iteration ranges and a schedule (OpenMP loop, GPU kernel, FPGA region replication, and others); and (3) in case control flow is necessary within a parallel region, SDFGs can be nested in each other with \textit{Nested SDFG} nodes.

\textbf{ONNX library nodes}\; The predefined semantics of library nodes allows them to be expanded to lower-level implementations, either ``native'' SDFG elements (tasklets, memlets, access nodes, and maps) or fast library calls (e.g., cuBLAS and cuDNN). As we shall show, native expansions can sometimes be optimized further than such libraries.
In DaCeML, each ONNX operator is represented by a library node. Native implementations for most ONNX operators use the DaCe NumPy frontend to generate SDFGs. For example:
\begin{minted}{python}
@python_pure_op_implementation
def Softplus(X, Y):
  Y[:] = numpy.log(1 + numpy.exp(X))
\end{minted}
Authoring operator implementations using this approach is simple and familiar to ML practitioners, and requires little regard for performance. These unoptimized implementations are later transformed using the recipe described in Section~\ref{sec:recipe}.

As not all operators are natively implemented, and the ONNX standard is expected to grow, we also support a fallback for any ONNX operator, using ONNX Runtime~\cite{onnxruntime} as a backend. Specifically, the expansions generate code that eagerly calls the operators within the compiled module.

\textbf{Progressive lowering of library nodes}\; Several library node expansions lower to other ONNX nodes. As an example, the \texttt{MatMul} node is lowered first to an Einstein sum (\texttt{Einsum}), which enables some tensor-contraction algebraic optimizations. Another is the \texttt{im2col} implementation of \texttt{Conv}, in which a \texttt{Gemm} node is included in the lowering output. This multi-level, progressive lowering reduces implementation redundancies and facilitates operator kernel authoring due to reuse.

\subsection{Symbolic Automatic Differentiation}

\begin{figure*}[t]
    \centering
    \includegraphics[width=0.3\linewidth]{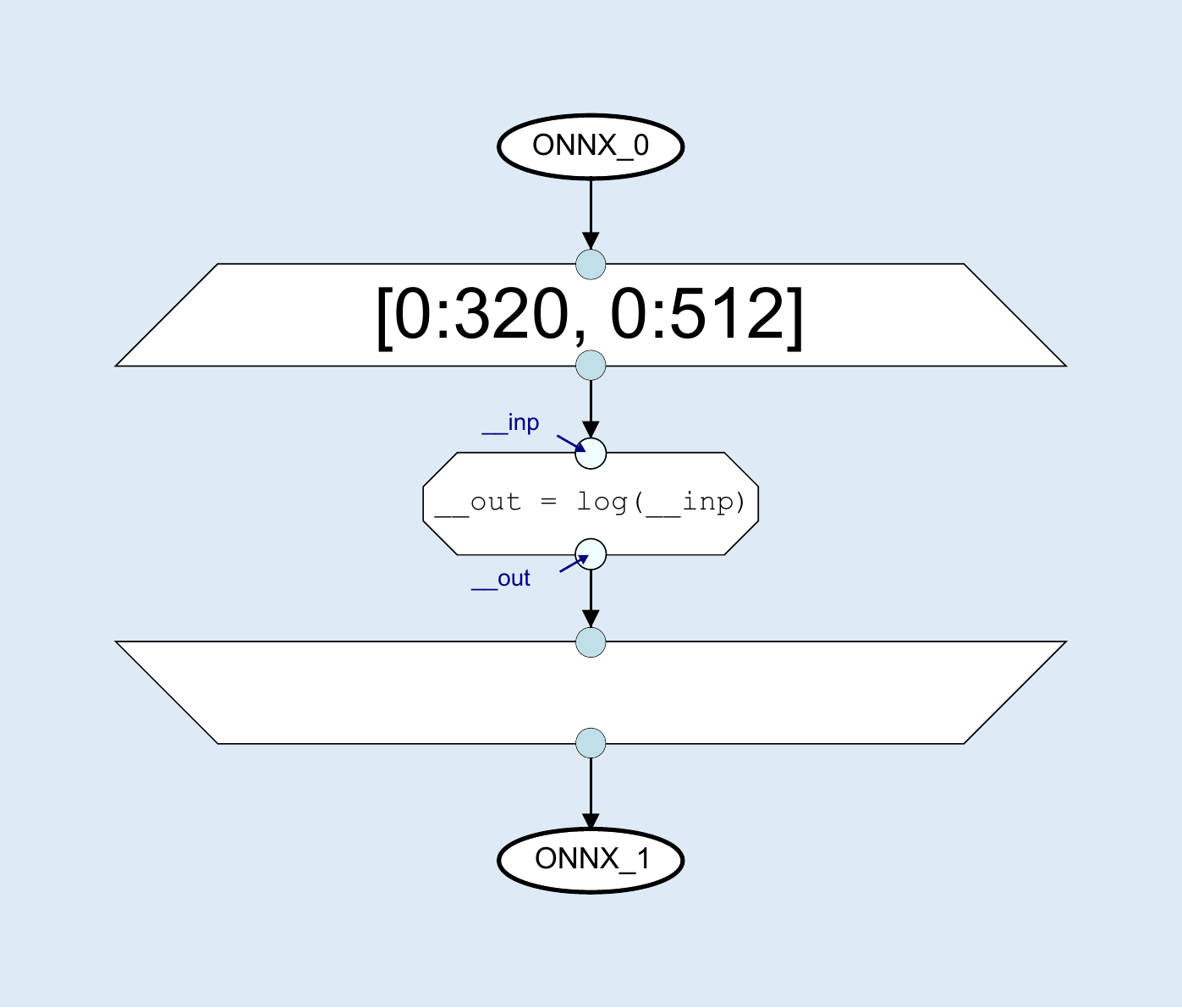}\qquad
    \includegraphics[width=0.3\linewidth]{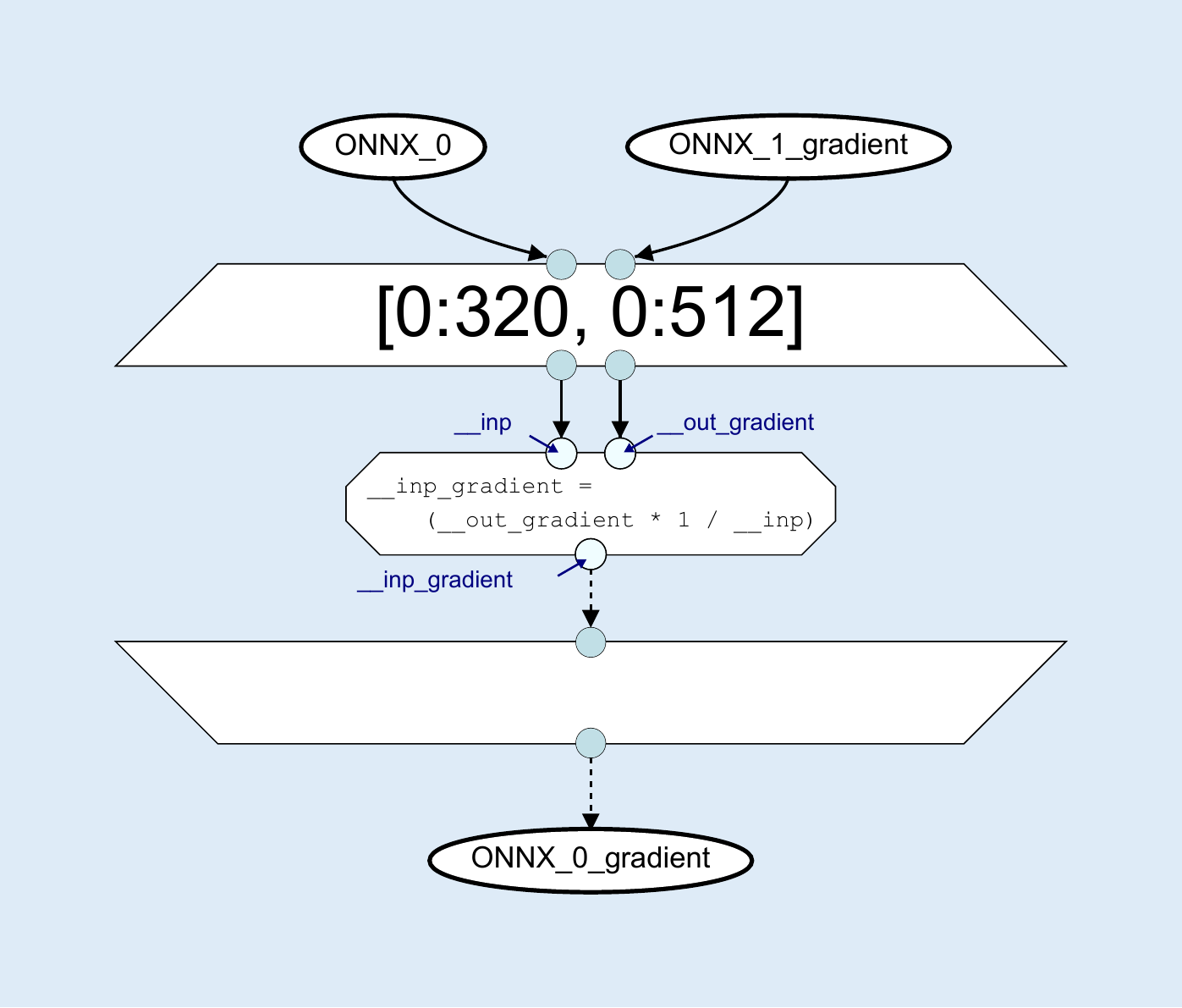}
    \vspace{-1em}
    \caption{Elementwise logarithm SDFG (left) and symbolic auto-differentiated version (right).}
    \label{fig:ad}
\end{figure*}
To compile and optimize for training, DaCeML uses source transformation based reverse-mode Automatic Differentiation (AD) to compute Vector-Jacobian Products (VJP) of the operators.
Guided by the principle of generality, all lower level optimization on the
backward pass is done using the same optimizing transformations as the
forward pass. To enable this, the DaCeML AD engine
transforms a forward pass SDFG into the corresponding SDFG that computes the required gradients. 
Critically, the differentiation is performed on the native SDFG, removing the need for manually specified VJPs. Furthermore, since all optimizations and lowering are performed on SDFGs, our AD engine is capable of differentiating scheduled and lowered SDFGs --- a capability that, to the best of our knowledge, no existing framework has.
By \emph{optimizing pre- and post-AD jointly}, we unlock a novel class of optimizations not expressible in other DNN frameworks.

Differentiating SDFGs is challenging, since nodes not only
represent computation, but memory accesses and parametric replication.
It is important to differentiate the computation symbolically, even if all sizes are known, since the access pattern (memlets) within each operator is still symbolic.

Given an SDFG, the engine first determines the subgraph to differentiate, based on the output
access nodes and the access nodes of inputs that require gradients (part of the metadata obtained from PyTorch). %
The subgraph
is then traversed in reverse topological order and reversed as follows. %
\textit{Access nodes} are reversed by “inverting” the access of the
node, and replacing the data that is written/read with the adjoint of
that data. 
\textit{Maps} can be reversed by simply converting map entry nodes to map
exit nodes and vice versa. 
The reverse of a code node --- whether \textit{tasklet, library node, or nested
SDFG} --- is a node that computes the VJP of the forward node. The inner tasklets of most machine learning operators work with scalar values. To automatically differentiate these scalar expressions, DaCeML uses the SymPy~\cite{sympy} symbolic differentiation engine (see example in Figure~\ref{fig:ad}).
Nested SDFGs are reversed recursively, and in the case of library nodes, they are lowered to their native SDFG implementation for further differentiation, with the possibility to provide manual backward expansions.

The VJPs of code nodes often require values that were inputs to the
corresponding forward node. These values are either \textit{forwarded} by storing the value in the forward pass and reading it in the backward pass, or recomputed in the backward pass. 

\begin{figure*}
    \centering
    \includegraphics[width=0.9\linewidth]{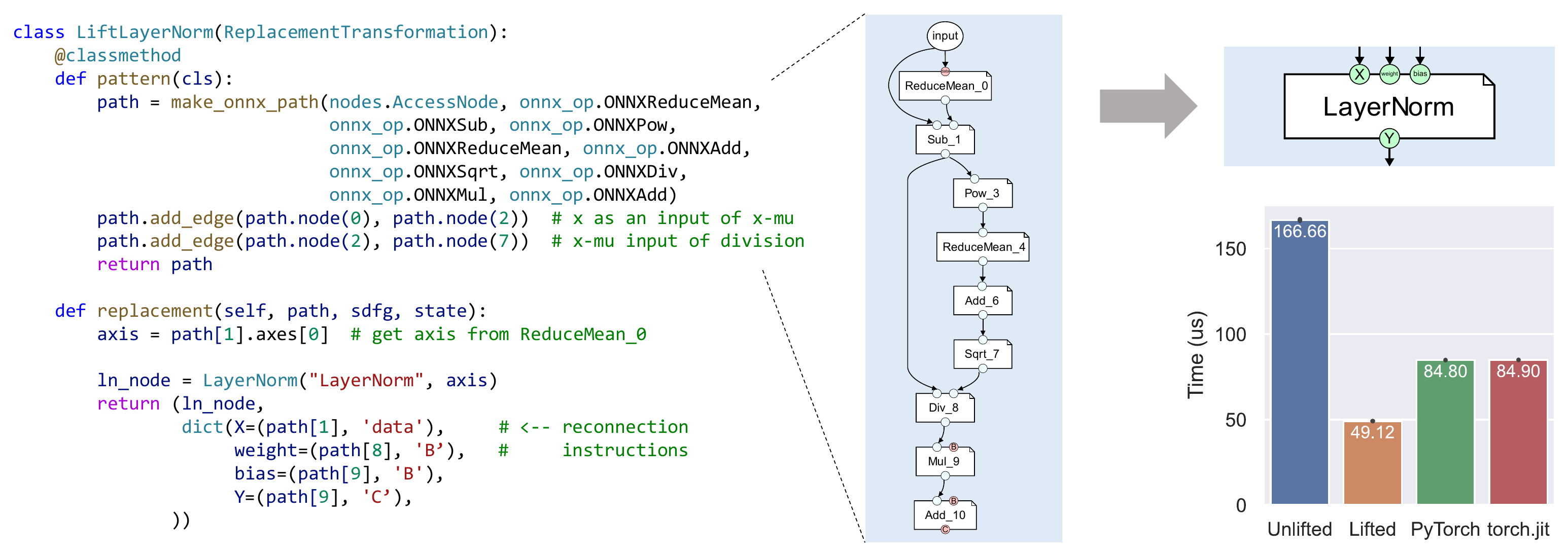}
    \vspace{-1em}
    \caption{Coarse-grained transformation example (Section~\ref{sec:pipeline:coarse}): lifting Layer Normalization from PyTorch.}
    \label{fig:liftln}
\end{figure*}

\textbf{Manual backward expansions}\; DaCeML's AD engine is able to automatically differentiate
almost all operator implementations, producing backward passes
comparable to the hand-written implementations. However, it is often beneficial to specify manual backward operators (e.g., \texttt{Einsum} can be reversed directly for performance or \texttt{Softmax}/\texttt{LogSoftmax} for numerical stability). DaCeML provides this capability by manual expansions.

\section{Optimization Pipeline}
\label{sec:pipeline}

DaCeML follows DaCe's white-box optimization approach, in which optimization can be performed automatically as a starting point, and then is interactive and guided by performance engineers. %
We further simplify the interface and extend its capabilities, in order to allow performance engineers and ML practitioners to productively write transformations of their own.

DaCeML transformations are written in Python, and can use the existing graph rewriting tools from DaCe. A transformation can perform local replacement (complete example in Figure~\ref{fig:liftln}), graph pattern-matching, or global modifications. The DaCeML replacement API is graph-based:  users provide the subgraph to match (\texttt{pattern}), and replacement instructions (\texttt{replacement}), which include returning a new node or subgraph and instructions on how to reconnect/augment the pattern graph (via the mapping in the return value). For more complex behavior, any method (exact matching condition, application, reconnection) can be overridden.

The transformations provided in DaCeML's standard set can be categorized into four distinct levels, ordered by their application as the graph is lowered: (1) Coarse-grained transformations on the ONNX-SDFG representation; (2) local data movement reduction, e.g., operator fusion and replication; (3) global (network-wide) data movement optimization; and (4) hardware specialization and workload partitioning. From those categories we sub-select transformations to apply globally on every network in our experiments, forming a ``recipe'' for DNN optimization, and show how guided application of the rest of them can yield even higher performance.

\subsection{Coarse-Grained Transformations}\label{sec:pipeline:coarse}

The first step in transforming the SDFG is leveraging the semantics of ONNX operators. We can further categorize this into transformations that clean up the graph from framework- and ONNX-related clutter, and transformations that find subgraph combinations to improve the performance or numerical stability of the model.

\textbf{Cleanup}\; We provide a general ONNX transformation called \texttt{ConstantFolding}, which evaluates nodes whose inputs are not part of the weights or model inputs, and replaces the subsequent paths with the evaluated value. Following expansion to native SDFG, another transformation (\texttt{InputToConstant}) complements this by inlining constant inputs into their respective tasklets.
This results in a chain of optimizations, for example, a 3.0 exponent in a \texttt{Pow} operator is first considered as a constant array; after cleanup, the code generator will subsequently replace the operation with much faster multiplications.
Overall, this eliminates many unnecessary copy and computational operations, and enables more complex transformations, such as algebraic fusion.

\textbf{Algebraic fusion}\; Due to the progressive lowering approach taken in DaCeML, many linear algebra operations (e.g., matrix-vector multiplication, transpose, batched tensor contraction) are lowered into \texttt{Einsum}s. We provide a set of transformations, such as horizontal and vertical  fusions, which can, for example, fuse together a transposition into a single Einsum (e.g., \texttt{ij->ji} and \texttt{ik,jk->ij} to \texttt{ki,jk->ij}). This information potentially feeds into combinatorial searches of optimal data layouts --- the shape of a weight can be modified through DaCeML if the Einsum does not generate an optimized BLAS call due to layouts (for which we provide a check). The ability to modify the shape of intermediates and weights is necessary to maximize performance in today's DNN architectures~\cite{ivanov2021data} and, to our knowledge, unique to DaCeML.

\textbf{Lifting and omitting operations}\; Other transformations relate to removing nodes that do not perform computations through symbolic size analysis, such as removing an Adaptive Pooling operator if the input/output sizes are the same, or fusing together padding and convolution operations. Known subgraphs generated by frameworks can be also be ``lifted'' to new nodes (e.g., with custom implementations or backward expansions), as is the case for Layer Normalization~\cite{layernorm} in Figure~\ref{fig:liftln}.

\subsection{Local Data Movement Reduction}

Domain-specific transformations on the ONNX representation do not suffice for optimization, as they only capture operator-centric behavior and will not work with unseen operators or unexpected combinations thereof.
We thus focus on transformations on the native SDFG (after ONNX library node expansion) that apply directly to ML workloads.

DaCe contains a library of standard transformations that use the structure of the graph to manipulate data movement. Of particular importance are \texttt{MapTiling} and \texttt{LocalStorage}, where the former partitions the workload of a map into tiles by introducing another nested map, and the latter adds an access node between two such maps in order to create storage only accessible by the current tile (e.g., placing weights in shared memory in Section~\ref{sec:effnet}). For DaCeML, we develop additional transformations to handle data movement bottlenecks in deep learning workloads.

\textbf{Transformations}\; Given an arbitrary subgraph, the \texttt{Subgraph\-Fusion} transformation tries to combine multiple map scopes into one scope. This reduces write/read roundtrips to global memory and, on GPUs, kernel launch overhead. In elementwise operations where iteration spaces are equal, the operation is trivial. However, in many cases the spaces are permuted (when different data layouts are involved) or do not share the entire iteration domain (e.g., in the Layer Normalization operator, see Section~\ref{sec:eval:layernorm}). For this reason, we extend subgraph fusion to find permutations, offsets (e.g., \texttt{start:end} vs. \texttt{0:end-start}), parallel regions in reduction, and the greatest common subset of map iteration domains to fuse over. The transformation extracts them out first and then fuses the requested outer maps, greatly extending the realm of fusion possibilities.

\textbf{Fusion space exploration}\; To automate the process of optimizing data movement in graphs, we develop automatic optimization heuristics that enumerate the space of fusible subgraphs. Currently, DaCeML supports greedy enumeration with pruning (based on path constraints for fusion), and ranking according to scoring functions. For evaluation, we rank by  largest neighboring regions to minimize data movement and GPU kernels.

\subsection{Backpropagation and Data Movement}

When optimizing the forward and backward passes at the same time, data-centric optimizations can control aspects of auto-differentia\-tion. 
As AD requires intermediate value forwarding, the memory footprint can become infeasible and movement too demanding~\cite{chen2016training,jain2019checkmate}.
However, fusing maps on the forward pass means omitting intermediate values, and replicating them means recomputation on the backward pass. This process can create a tunable ``knob'' to optimize backpropagation.

For this purpose, we develop two transformations: \texttt{Tasklet\-Fusion}, which fuses two computations into one symbolic tasklet; and \texttt{Map\-Replication}, a transformation that detects access nodes being read more than once, and replicates the immediate map leading to it. %
Example uses of the two on the Mish activation~\cite{mish} and statistical normalization are shown in Section~\ref{sec:eval}.

\subsection{Global Data Layout Optimizations}

After tuning local data movement, one can look beyond the traditional ``peephole optimizations'' and schedule data assignment and movement on the entire graph with data-centric transformations.
Briefly, such transformations do not have a pattern to match, but view the entire graph and generate multiple options. One such optimization is \texttt{TransientReuse}, which detects arrays that have the same volume but are used in non-overlapping segments of time (using DAG level analysis). References to these arrays are replaced by a reference to a single memory region, and unused arrays are removed. Another potential example is finding an optimal set of data layouts for weights and intermediate values. Since DaCeML generates native SDFGs, it can generate optimized code for each layout, constrain the search space by ensuring layouts match, and prune it based on domain-specific constraints, e.g., by only considering BLAS-optimized layouts for Einsums.

Allocation and deallocation of memory can create significant overheads when happening within critical code. To avoid them, the SDFG provides fine-grained control over their lifetimes. For example, when storage is annotated as Persistent, it will outlive multiple SDFG invocations.    

\subsection{Hardware Specialization}
\label{sec:hardware}
Lastly, we need to consider the underlying system we are compiling for. This may mean the accelerator architecture(s), or whether we are running on one or more nodes.

\textbf{GPU specialization}\; Data-centric transformations can offload code to different platforms by introducing copies and kernel code~\cite{ben2019stateful}.
In DaCeML, we extend the transformation capabilities to partition workload efficiently on GPUs. \texttt{WarpTiling} is a variant of \texttt{MapTiling} that takes a GPU kernel map and divides its work across a warp. It detects reductions inside the map and inserts efficient warp-collective reductions as necessary.
\texttt{Vectorization} allows using target-specific vector instructions. We extend the system to support reduced-precision vector types, provide fast implementations of vector-to-scalar reductions, and ``fill in gaps'' in math library functions by calling sequences of scalar operations.

\textbf{Distributed computing}\; DaCeML supports data-parallelism for distributed training.  \texttt{DistDataParallel} adds a distributed \texttt{all\-reduce} library node after each weight gradient access node in the backward-pass SDFG. Since the code generator traverses SDFGs in topological order, communication is performed as soon as the data are ready, promoting pipelining.

\subsection{Automatic Transformations}\label{sec:recipe}
Now that we have established a set of transformations, we can create an automated sequence of their application for arbitrary DNNs. Our transformation recipe used in the results is as follows: (1) Clean up the ONNX-based SDFG with constant folding and algebraic fusion; (2) lower ONNX nodes to implementations, choosing from ONNXRuntime, cuDNN, or PyTorch implementations depending on the node type; (3) in the resulting SDFG, inline nested SDFGs, remove all redundant copies, and greedily fuse subgraphs; (4) specialize the program to the hardware: in fused maps, hoist out initialization to kernel start, eliminate trivial (0, 1 element) maps, specialize memory types of transients and call \texttt{WarpTiling}; (5) flatten map dimensions to coalesce memory accesses, and vectorize maps to 128-bit accesses if applicable. When manually applying guided transformations, this recipe can be used as a starting point.

\section{Evaluation}
\label{sec:eval}
We first demonstrate DaCeML's ability to optimize single operators using the recipe described in Section~\ref{sec:recipe}, and then examine how these perform when used as part of larger models.

\textbf{Experimental setup}\; We run each experiment at least 100 times and report the median value with a 95\% nonparametric confidence interval. %
For measurement, we use a server with an Intel Xeon Gold 6140 CPU (2.30GHz), 768 GiB RAM, and an NVIDIA Tesla V100 GPU (16 GiB RAM). We use Python 3.8.8, DaCe 0.10.8, CUDA 11.4, cuDNN 8.2.0.54. PyTorch 1.8.1, ONNXRuntime 1.7.0, TensorFlow 2.5.0, TensorRT 8.0.1.6, TVM 0.8.0dev0 (commit dbfbebe), Triton 1.1.1, and JAX 0.2.13.
The torch-based frameworks were all run from the same model source code, unmodified except for the addition of a decorator for \texttt{torch.jit} and DaCeML.

\subsection{Mish Activation}\label{sec:eval:mish}
\begin{figure}
    \centering
    \begin{subfigure}[b]{\linewidth}
    \centering
    \includegraphics[width=0.4\linewidth]{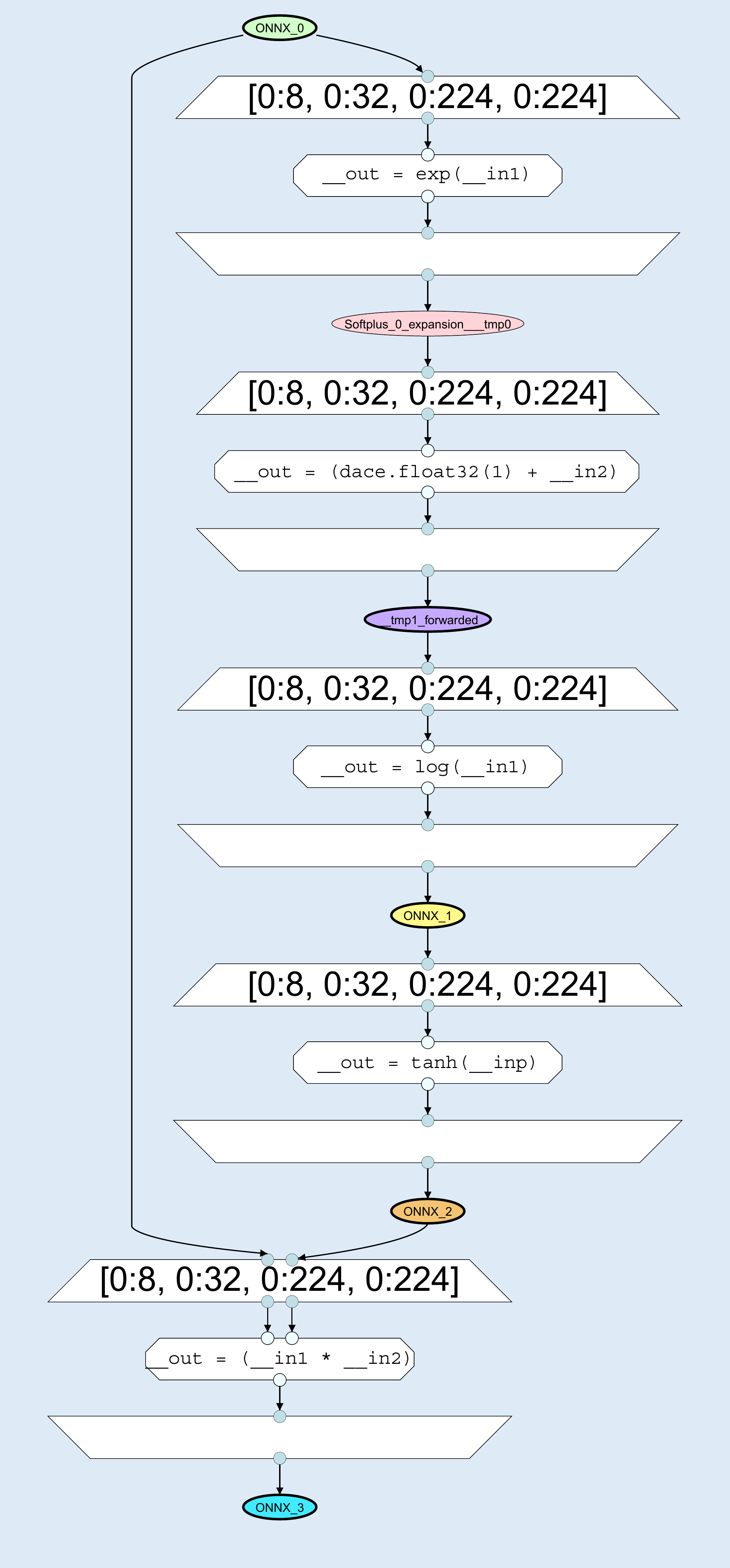}
    \quad
    \includegraphics[width=0.45\linewidth]{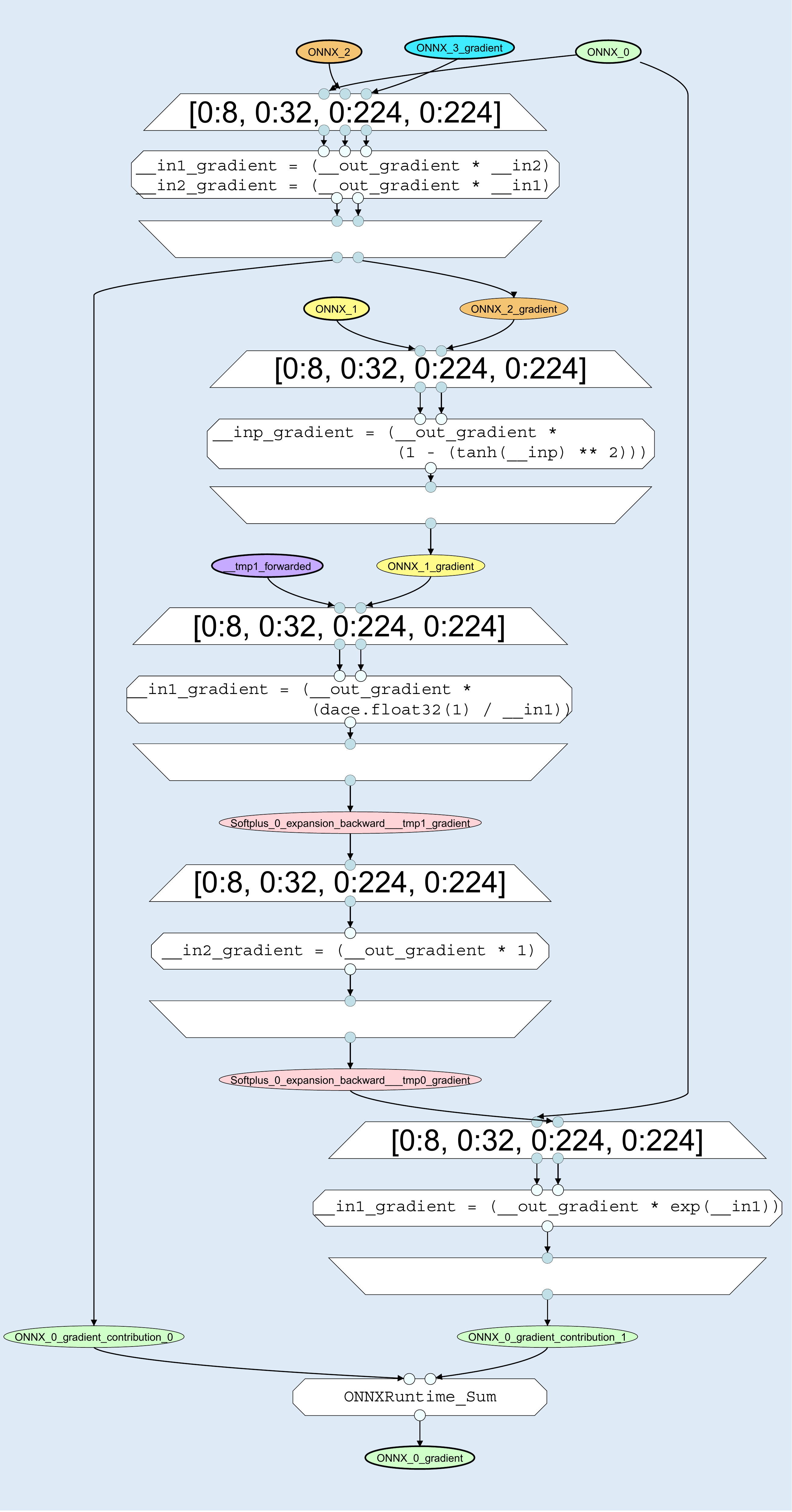}
    \vspace{-0.5em}\caption{Before Optimization (intermediate memory: 134.75MiB)}\vspace{0.5em}
    \label{fig:mish:before}
    \end{subfigure}
    \begin{subfigure}[b]{\linewidth}
    \centering
    \includegraphics[width=0.425\linewidth]{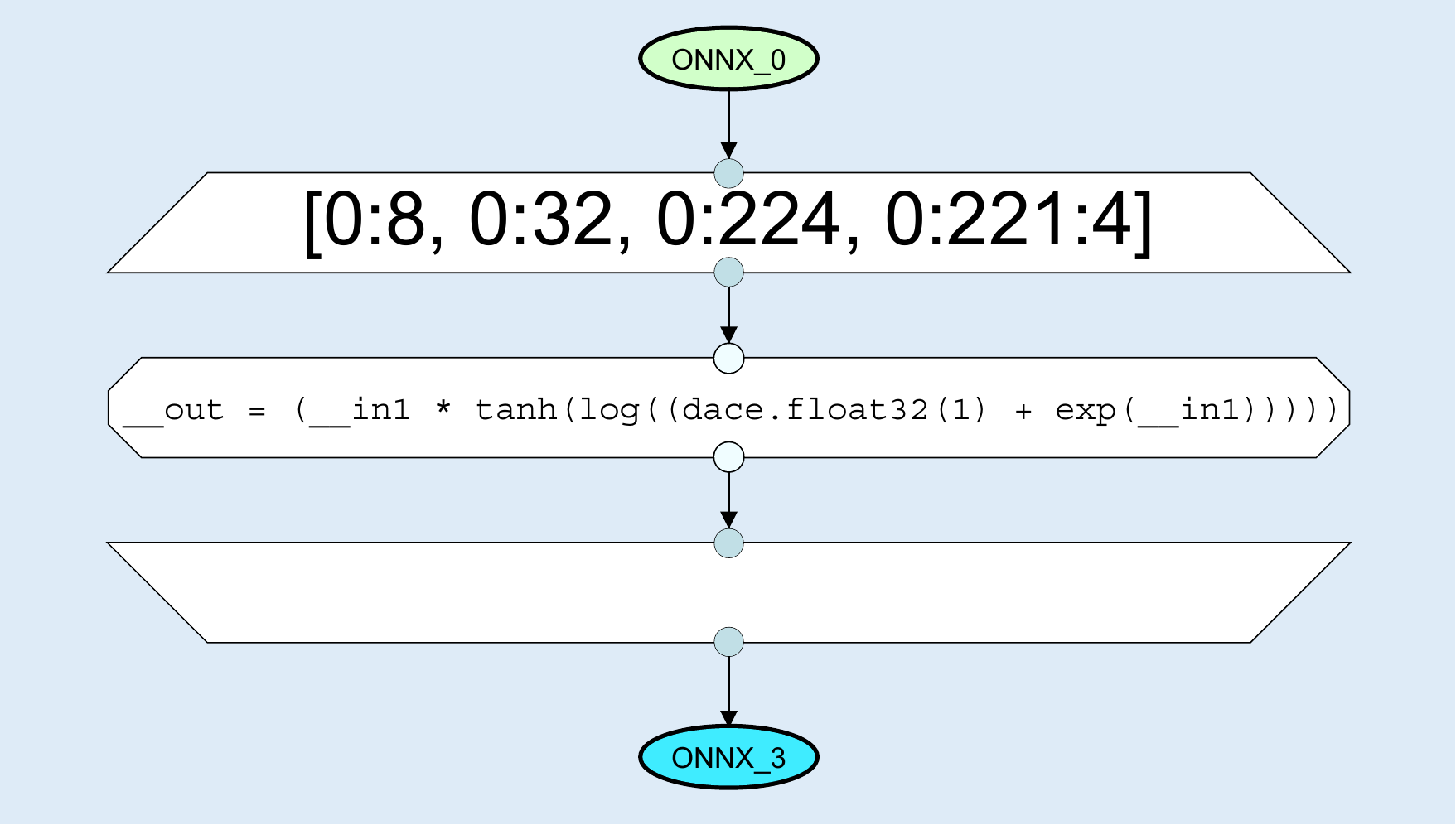}
    \quad
    \includegraphics[width=0.425\linewidth]{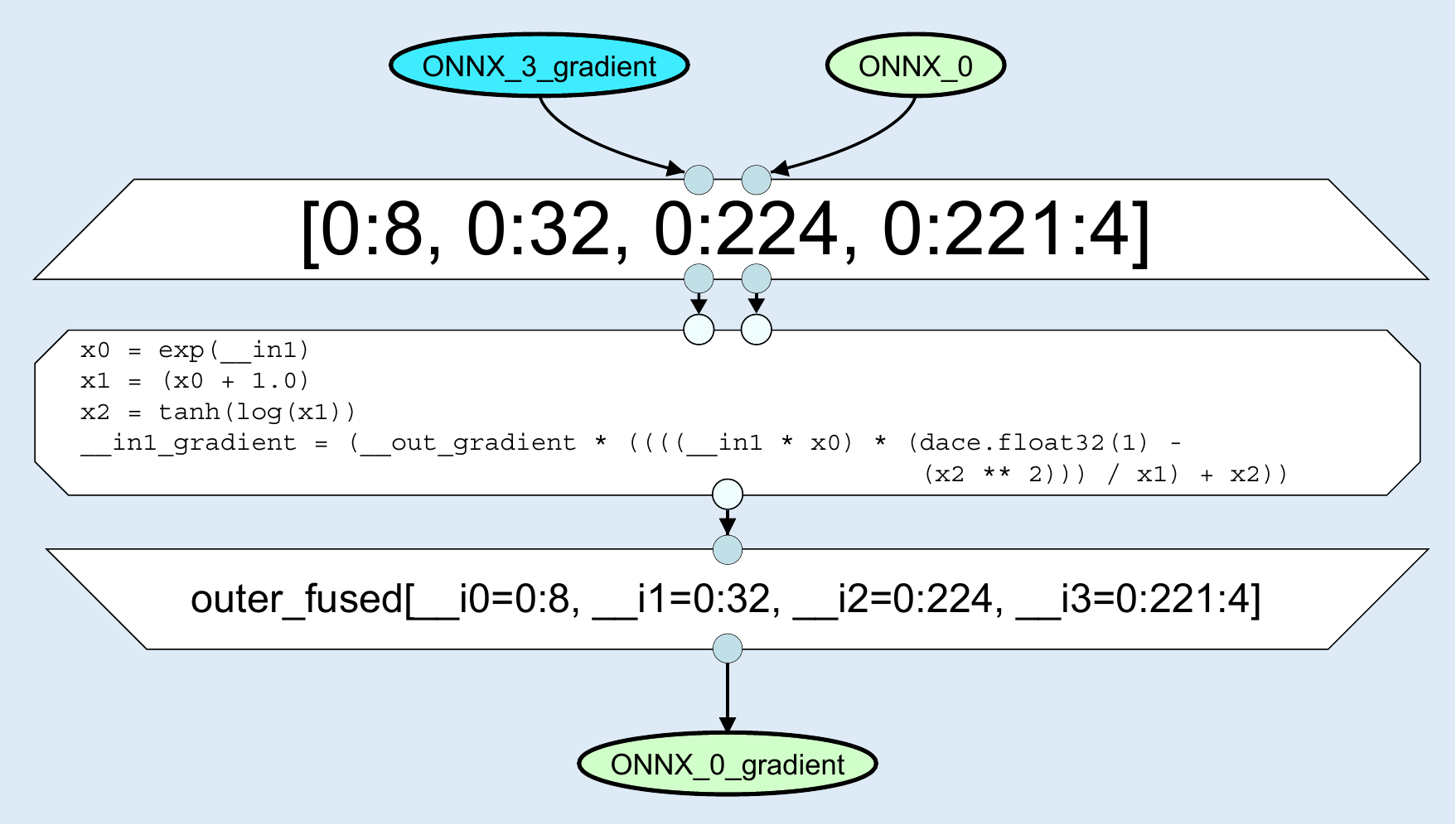}
    \vspace{-0.5em}\caption{After Optimization (intermediate memory: 0MiB)}
    \label{fig:mish:after}
    \end{subfigure}
    \caption{Mish operator SDFG forward pass (left) and the resulting backpropagated graph (right). Colored access nodes correspond to matching data.}
\end{figure}

\begin{figure*}
    \centering
    \begin{subfigure}[b]{0.25\linewidth}
\begin{minted}[fontsize=\scriptsize, escapeinside=||]{python}
@op_implementation(op="Softmax", name="native")
class NativeSoftmax(ONNXForward):
  @staticmethod
  def forward(node, state, sdfg):
    # Get property into scope
    axis = node.axis

    # Operator implementation in numpy
    def prog(input, output):
      |\colorbox[HTML]{D1FFC9}{max = input.max(axis=axis, keepdims=True)}|
      |\colorbox[HTML]{FFD4D8}{exp = np.exp(input - max)}|
      |\colorbox[HTML]{C6AAFF}{sum = exp.sum(axis=axis, keepdims=True)}|
      |\colorbox[HTML]{3FECFF}{output[:] = exp / sum}|

    return program_for_node(prog, sdfg,
                            state, node)



||
\end{minted}
    \caption{NumPy Implementation}
    \label{fig:smax:code}
    \end{subfigure}\quad
    \begin{subfigure}[b]{0.2\linewidth}
    \centering
    \includegraphics[height=2.5in,page=1]{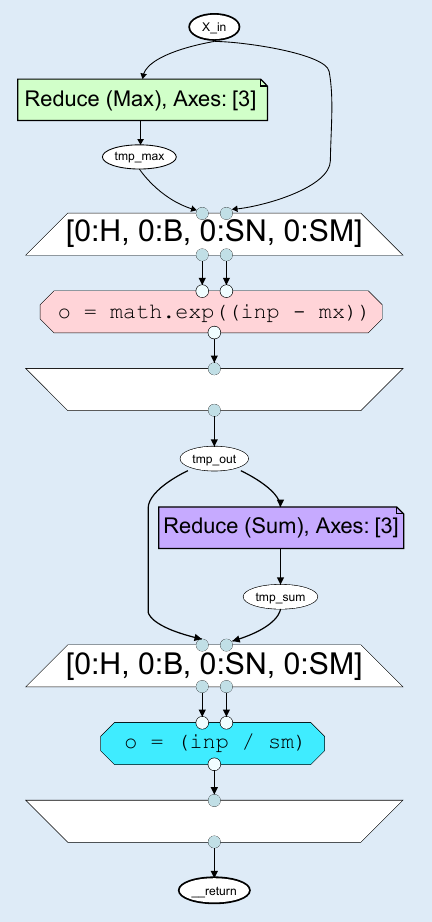}
    \caption{Initial SDFG}
    \label{fig:smax:initial}
    \end{subfigure}\quad
    \begin{subfigure}[b]{0.2\linewidth}
    \centering
    \includegraphics[height=2.5in,page=2]{figures/smax1.pdf}
    \caption{Subgraph Fusion}
    \label{fig:smax:sgf}
    \end{subfigure}
    \begin{subfigure}[b]{0.15\linewidth}
    \centering
    \includegraphics[height=2.5in,page=3]{figures/smax1.pdf}
    \caption{Specialization}
    \label{fig:smax:spec}
    \end{subfigure}
    \begin{subfigure}[b]{0.15\linewidth}
    \centering
    \includegraphics[height=2.5in,page=4]{figures/smax1.pdf}
    \caption{Final}
    \label{fig:smax:final}
    \end{subfigure}\vspace{-1em}
    \caption{Softmax implementation in DaCeML (colored sections correspond to same operators).}
    \label{fig:smax}
    \vspace{-1em}
\end{figure*}

\begin{figure}[t]
    \centering
    \includegraphics[width=0.48\linewidth]{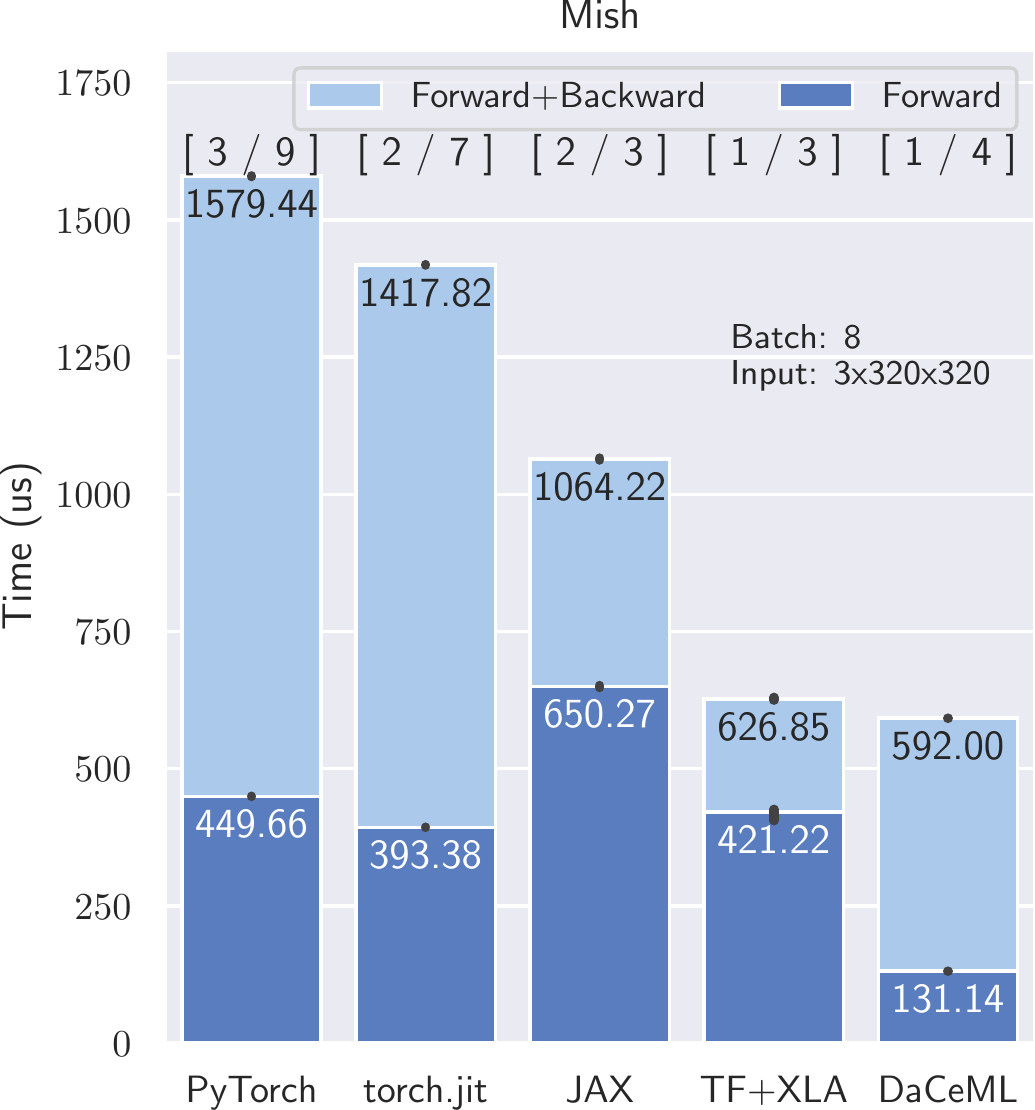}
    \hfill
    \includegraphics[width=0.48\linewidth]{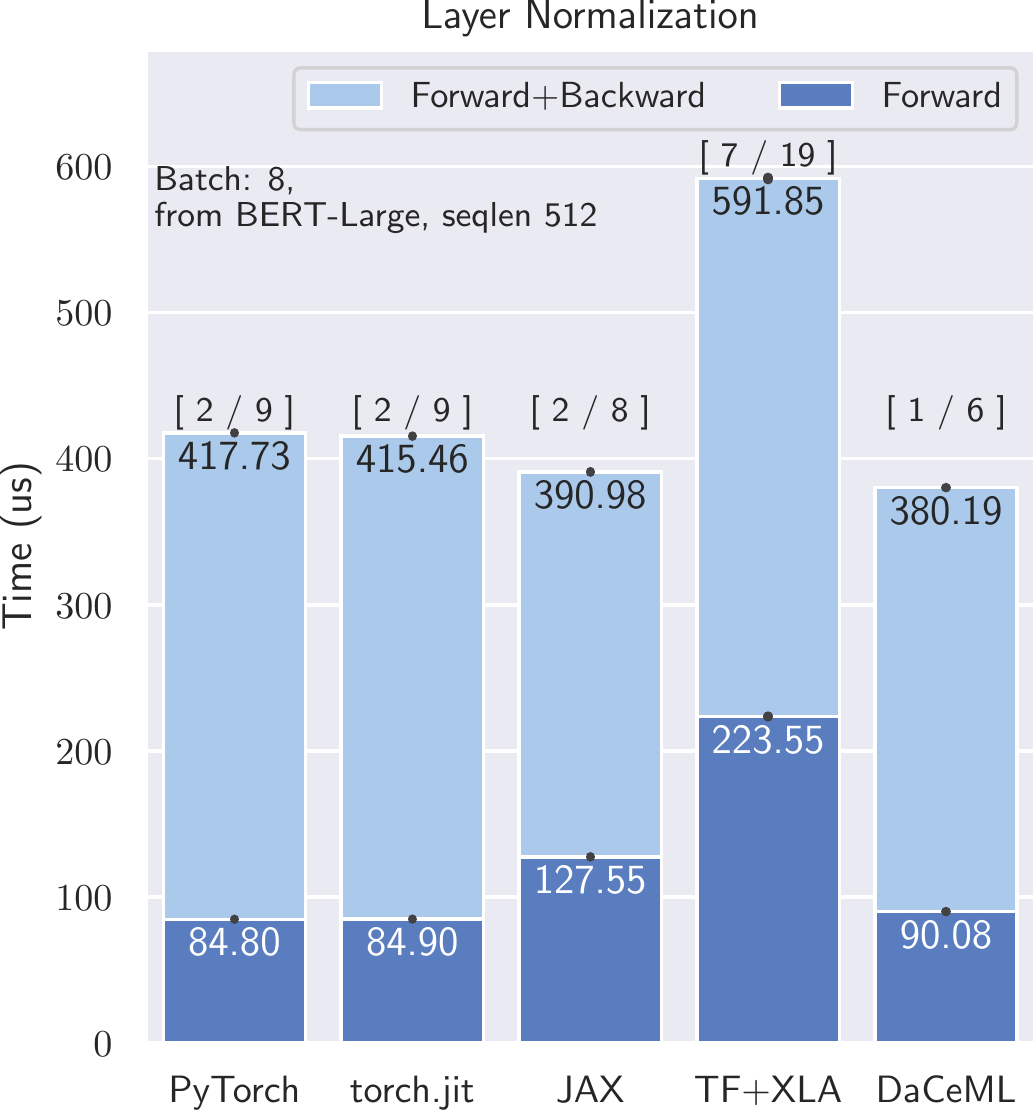}
    \caption{Runtime results for DNN operators. Brackets $[ \cdot ]$ indicate number of kernels launched in forward / fwd+bwd.}
    \label{fig:eval-blocks}
    \vspace{-1em}
\end{figure}

YOLOv4~\cite{yolov4} is a widely-used network for object detection.
Among other features, it relies on the Mish~\cite{mish} activation: $\mathrm{mish}(x) = x\tanh(\mathrm{softplus}(x))$. As with many operators, Mish is not natively supported by ONNX or ONNX Runtime.

Running Mish for training in PyTorch yields the graph shown in Figure~\ref{fig:mish:before}. The operator consists of sequential elementwise application of the individual primitives used in the function (i.e., $\exp$, $\tanh$, $\log$, $+$, $*$), with intermediate arrays generated to be retained for the backward pass, as can be seen on the right-hand side of the figure. These arrays are global and cannot be optimized once the AD process is complete, which is when they would typically be optimized on most DNN compilers.

The DaCeML generic transformation recipe lives both before and after AD, and yields an optimized SDFG prior to AD as shown in Figure~\ref{fig:mish:after} (left). As the kernels are entirely fused, the decision is made to recompute intermediate values for differentiation on a scalar level, rather than stash them for the backward pass. Differentiating the fused and scheduled kernel, a capability that current deep learning frameworks lack, produces a backpropagated graph that minimizes data movement, as shown in the figure (right). DaCeML is currently, to the best of our knowledge, \emph{the only deep learning framework with this capability}.

We evaluate Mish in isolation in Figure~\ref{fig:eval-blocks} (left).
By fully fusing the graph \textit{before} differentiation, DaCeML improves performance over all other tested frameworks.
For forward evaluation, PyTorch produces three kernels: $\tanh$, $\mathrm{softplus}$, and multiplication.
\texttt{torch.jit} and JAX are able to fuse two out of the three operations.
TensorFlow+XLA manages to produce a single fused kernel but fails to eliminate global loads/stores that stash intermediate values for the backward pass. 
DaCeML achieves $3.43\times$ improvements over PyTorch and $3\times$ over \texttt{torch.jit}.
In backpropagation, we similarly achieve $2.67\times$ improvements over PyTorch and $1.06\times$ over TF+XLA.

\subsection{Statistical Normalization}\label{sec:eval:layernorm}
Statistical normalization operations (e.g., Batch Normalization~\cite{ioffe2015batch}, Lay\-er Normalization~\cite{layernorm}, Group Normalization~\cite{wu2018group}) are commonplace in modern DNNs. 
Despite the fact that all normalization operations are similar, their performance varies. All the aforementioned operations perform the general computation $\gamma\cdot\frac{x-\mathbb{E}[x]}{\sqrt{\text{Var}(x) + \epsilon}}+\beta$, but on different dimensions or subsets thereof. 
We examine layer normalization~\cite{layernorm}, a primitive widely used in transformer models.

We find that PyTorch has two separate implementations for layer and batch normalization.
The implementation of batch normalization can use cuDNN, while layer normalization uses a manually-optimized function. Thus, despite batch normalization being more expensive than layer normalization (due to saving running mean/variance statistics to memory), the former is faster than the latter.

We evaluate the implementation of layer normalization following the DaCeML recipe, listing its performance in Figure~\ref{fig:eval-blocks} (right).
The generated code outperforms all frameworks on the forward-and-backward pass, where JAX closely matches in backpropagation but misses several fusion opportunities. On the forward part, PyTorch is slightly faster due to DaCeML storing more information for backpropagation, which ends up being faster overall. Without this storage, DaCeML takes 49 $\mu$s (Figure~\ref{fig:liftln}). DaCeML's native SDFG is also fusible with neighboring operators, which delivers improved overall performance when used in larger models.

\subsubsection{Softmax}
This operator is present in many networks, both for classification and as a key component of attention mechanisms~\cite{vaswani2017attention}.
Even though it is computationally different from normalization, it exhibits a similar data movement pattern. Thus, the same data-centric transformations performed on layer normalization apply to softmax directly. Below, we dive into the transformations applied, showing how the representation and new DaCeML transformations can reduce data movement in nontrivial ways, regardless of the underlying computation.

The input NumPy implementation of the operator (Figure~\ref{fig:smax:code}) %
lowers the ONNX library node to four separate operations: two reductions (computing the maximum value to subtract; and summarizing the contributions for the denominator), and two elementwise operations (exponentiation, division).
In the initial graph (Figure~\ref{fig:smax:initial}), three extraneous tensors sized as the input would be generated. The second step of our recipe locally reduces this data movement.

Fusion of these operators is not trivial, as the dimensions of the maps and reductions differ. The greedy fusion strategy applied in DaCeML automatically expands reductions and finds common dimensions to extract out of the maps in the subgraph, which in turn allows to fuse them all to one map with the range \texttt{0:H, 0:B, 0:SN}. This stores all memory locally (within the map) and will be allocated as GPU registers upon code generation (Figure~\ref{fig:smax:sgf}). As the global data movement optimization step does not apply in this case, the recipe proceeds to specialize the code to the GPU. 

The optimization pipeline gradually lowers the representation, and applies the DaCeML \texttt{WarpTiling} transformation to partition the work in the internal maps across warps (Figure~\ref{fig:smax:spec}), and automatically generates warp-level reductions for max/sum. To avoid extra data allocation, the recipe \emph{replicates} all computations that have multiple results dependent on reductions by default, minimizing data movement at the cost of redundant computations.
Lastly, the SDFG is cleaned up via another pass of transformations --- front-loading reduction initialization, map fusion, and vectorization (Figure~\ref{fig:smax:final}).

\begin{figure}[t]
    \centering
    \includegraphics[height=1.7in]{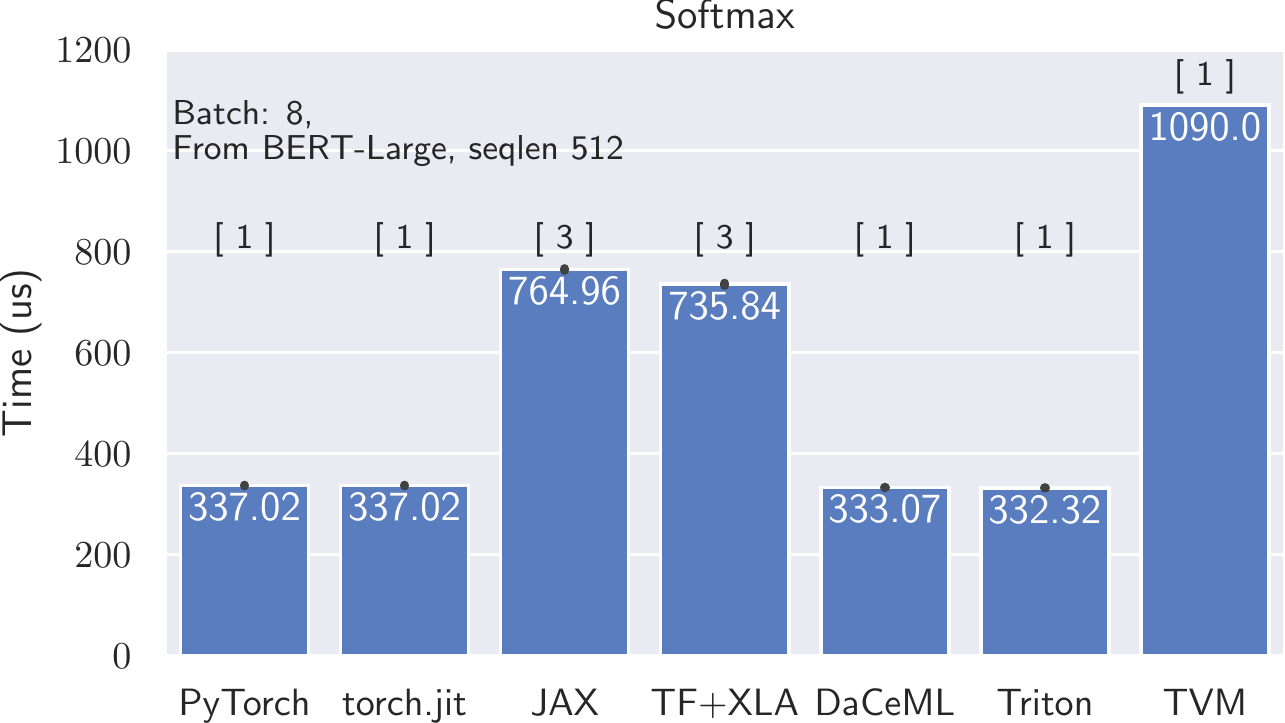}
    \caption{Runtime results for the softmax operator. Brackets $[ \cdot ]$ indicate number of kernels launched.}
    \label{fig:eval-smax}
    \vspace{-1em}
\end{figure}

Due to the popularity of the operator, many frameworks include optimized implementations of softmax. In Figure~\ref{fig:eval-smax}, we evaluate our generated implementation against several competing frameworks, including ones that do not provide learning capabilities. Of the tested frameworks, Triton~\cite{tillet2019triton} (which uses a hand-written kernel) and DaCeML achieve the best performance. The hand-written kernel included in PyTorch hardcodes a similar warp-tiling scheme as the DaCeML pipeline optimizes to, yet runs slightly slower. The XLA-based frameworks fail to fully fuse the operator, resulting in 3 kernels. While TVM contains a Relay operator for softmax, it is not tunable with AutoTVM, and does not contain warp-level primitives, both of which contribute to its high runtime relative to the warp-optimized frameworks. As detailed in Section~\ref{sec:eval:bert}, the DaCeML implementation is fully fusible with neighboring operations, and applying the recipe in the BERT transformer DNN results exploits this for an additional 196 $\mu s$ gain per call. Handwritten kernel implementations of other frameworks (e.g., Triton, PyTorch) are generally not fusible.

\subsection{Inference with Optimized Operators}
\begin{figure}[t]
  \centering
  \includegraphics[height=1.7in]{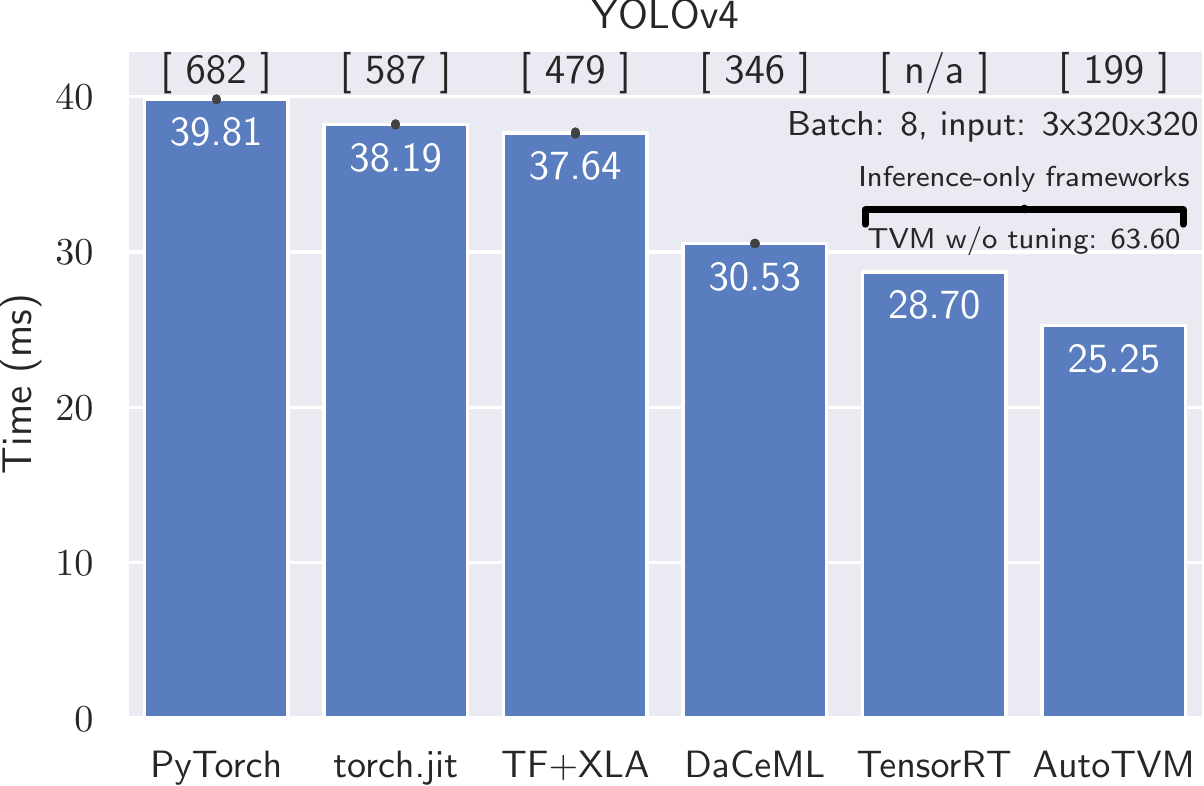}
  \caption{Runtime results for inference with YOLOv4. Brackets $[ \cdot ]$ indicate number of kernels launched. Due to profiler issues, kernel counts for TensorRT could not be obtained.}
  \label{fig:eval-yolo}
  \vspace{-1em}
\end{figure}

Figure~\ref{fig:eval-yolo} lists inference results for the YOLOv4 DNN with the automated recipe, which uses the optimized Mish activation (Section~\ref{sec:eval:mish}) in the context of the full network. The figure shows a reduction in kernel counts, as well as 1.22--1.3$\times$ speedup over training frameworks. This nearly reaches the optimization levels of inference-only frameworks, such as TensorRT and TVM. Upon deeper inspection, the two frameworks make use of inference-specific optimizations, such as custom implementations for convolutions and pre-transforming weights. As DaCeML is designed to optimize training workloads, such techniques are outside the scope of this work.

\subsection{Automatic Optimization for Training}
\begin{table*}
  \small
  \centering
  \begin{tabular}{llc@{\hspace{2mm}}cc@{\hspace{2mm}}cc@{\hspace{2mm}}cc@{\hspace{2mm}}cc@{\hspace{2mm}}c}
    \toprule
    & & \multicolumn{2}{c}{PyTorch}           & \multicolumn{2}{c}{torch.jit}     & \multicolumn{2}{c}{JAX}           & \multicolumn{2}{c}{TF+XLA}        & \multicolumn{2}{c}{DaCeML}        \\
    & & $\rightarrow$ & $\rightleftarrows$    & $\rightarrow$ & $\rightleftarrows$ & $\rightarrow$ & $\rightleftarrows$ & $\rightarrow$ & $\rightleftarrows$ & $\rightarrow$ & $\rightleftarrows$ \\
    \midrule\\\addlinespace[-0.75em]
    \hspace{-2mm}\multirow{2}{*}{\rotatebox{90}{Guided}}& &&&&&&&&&&\\\addlinespace[-0.75em]
    & \faEye~EfficientNet~\cite{tan2019efficientnet}
    & 2.05          &  6.90                & 2.04          & 6.94             & 2.39          & 7.40             & 1.54          & 6.37             & \textbf{1.40}          & \textbf{5.97}             \\
    & \faRobot~BERT$_{\textsc{LARGE}}$ {\tiny(mixed)}~\cite{devlin2019bert}
    & 2.94          &  8.18                & 2.92          & 8.20             & 3.19          & 8.11             & 3.80          & 10.76             & \textbf{2.74}          & \textbf{7.62}             \\\addlinespace[0.5em]
    \midrule
    \hspace{-2mm}\multirow{10}{*}{\rotatebox{90}{Automatic}}
    & \faEye~ResNet-50~\cite{he2015deep}
    & 14.55          &  32.04                & \textbf{9.98}          & \textbf{31.94}             & 14.17          & 33.93             & 12.33          & 35.57             & 10.03          & 32.45             \\
    & \faEye~Wide ResNet-50-2~\cite{zagoruyko2017wide}
    & 22.50          &  70.94                & 22.45          & 70.83             & 40.49          & 98.13             & 32.79          & 99.06             & \textbf{20.62}          & \textbf{67.99}             \\
    & \faEye~MobileNet V2~\cite{sandler2019mobilenetv2}
    & 9.98          &  18.45                & 6.22          & 15.53             & ---          & ---             & 7.42          & 20.29             & \textbf{4.74}          & \textbf{14.77}             \\
    & \faEye~EfficientNet~\cite{tan2019efficientnet}
    & 2.05          &  6.90                & 2.04          & 6.94             & 2.39          & 7.40             & \textbf{1.54}          & \textbf{6.37}             & 1.57          & 15.00             \\
    & \faLowVision~MLP Mixer~\cite{tolstikhin2021mlpmixer}
    & 1.63          &  \textbf{3.65}                & \textbf{1.36}          & 3.66             & 1.77          & 4.01             & ---          & ---             & 1.48          & 4.25             \\
    & \faPuzzlePiece~FCN8s~\cite{long2015fcn}
    & 46.85          &  158.42                & 46.82          & \textbf{158.40}             & ---          & ---             & ---          & ---             & \textbf{45.97}          & 166.30             \\
    & \faVolumeUp~WaveNet~\cite{oord2016wavenet}
    & 23.21          &  46.39                & \textbf{18.67}          & 41.49             & ---          & ---             & ---          & ---             & 26.16          & \textbf{41.07}             \\
    & \faRobot~BERT$_{\textsc{LARGE}}$ {\tiny(single)}~\cite{devlin2019bert}
    & 11.05          &  31.76                & 11.05          & 31.82             & \textbf{10.93}          & \textbf{29.94}             & 11.14          & 38.73             & 11.44          & 32.98             \\
    & \faRobot~BERT$_{\textsc{LARGE}}$ {\tiny(mixed)}~\cite{devlin2019bert}
    & 2.94          & 8.18                & \textbf{2.92}          & 8.20             & 3.19          & \textbf{8.11}             & 3.80          & 10.76             & 3.34          & 9.25             \\
    & \faChartLine~DLRM~\cite{naumov2019dlrm}
    & 118.07          &  126.55                & \textbf{117.38}          & 126.83             & ---          & ---             & ---          & ---             & 117.69          & \textbf{126.42}             \\
    \bottomrule
\end{tabular}
\caption{Median runtime in milliseconds of the forward ($\rightarrow$) and forward + backward ($\rightleftarrows$) passes for convolutional vision (\faEye), non-convolutional vision (\faLowVision), audio (\faVolumeUp), image segmentation (\faPuzzlePiece), transformer (\faRobot) and recommendation system (\faChartLine) models. BERT$_{\textsc{LARGE}}$ corresponds to a single encoder layer; EfficientNet to the first MBConv layer. A --- indicates an implementation was not found for the model.}
\label{tab:automatic}
\end{table*}
While DaCeML's major strength lies in its manual tuning capabilities, the automated transformation recipe from Section~\ref{sec:recipe} already yields performance improvements over state-of-the-art frameworks. We demonstrate networks from different domains, with varying data movement patterns, as well as utilizing different hardware units (e.g., tensor cores). We run the following models with the recipe alone, and list the results in Table~\ref{tab:automatic} (bottom): ResNet-50~\cite{he2015deep}, Wide ResNet-50-2~\cite{zagoruyko2017wide}, MobileNet V2~\cite{sandler2019mobilenetv2}, EfficientNet-B0's MBConv block~\cite{tan2019efficientnet}, MLP Mixer~\cite{tolstikhin2021mlpmixer}, Fully Convolutional Network~\cite{long2015fcn}, WaveNet~\cite{oord2016wavenet}, BERT$_{\textsc{LARGE}}$~\cite{devlin2019bert} encoder block in single (32-bit) and mixed (16-bit) precision, and the DLRM~\cite{naumov2019dlrm} recommendation system. An interested ML practitioner could then use DaCeML to further optimize performance as necessary.

The table shows that no single framework operates best across all DNNs. Additionally, with the automatic optimizations alone, we see that DaCeML roughly matches and in multiple cases outperforms the other compiler frameworks. This is especially interesting in variants of popular networks, such as Wide ResNets. DaCeML is roughly 2$\times$ slower on Wide ResNet-50-2 than on ResNet-50, as expected for performing twice the computations; yet other frameworks are up to 2.89$\times$ slower. This potentially indicates specialization for certain operators and sizes, which does not occur with the data-centric transformations in DaCeML.

We now proceed with two case studies that highlight the possibilities enabled by DaCeML's user-guided optimization workflow, where the results are summarized in Table~\ref{tab:automatic} (top).

\subsection{Guided Optimization Case Study: EfficientNet}\label{sec:effnet}
\begin{figure}
    \centering
    \includegraphics[width=0.48\linewidth]{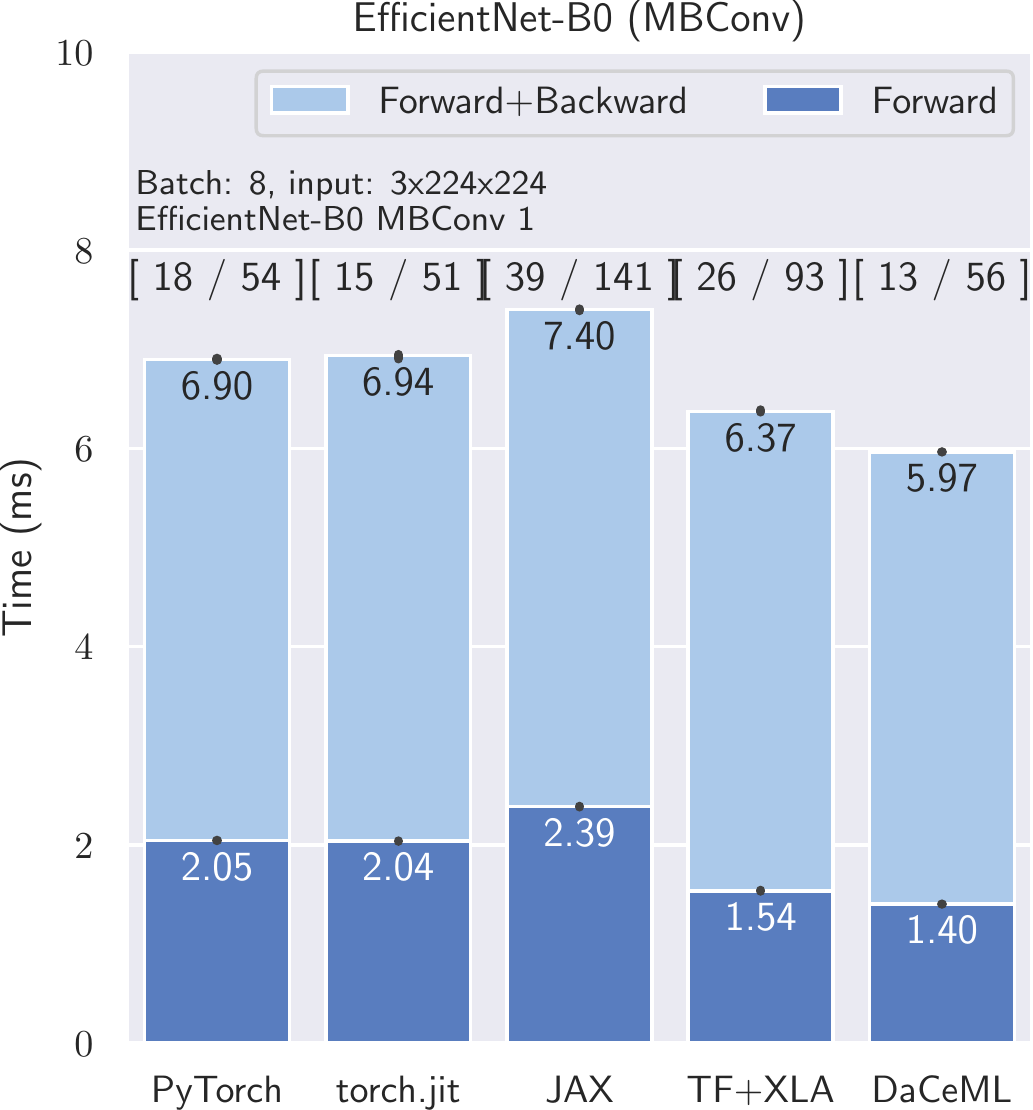}
    \hfill
    \includegraphics[width=0.48\linewidth]{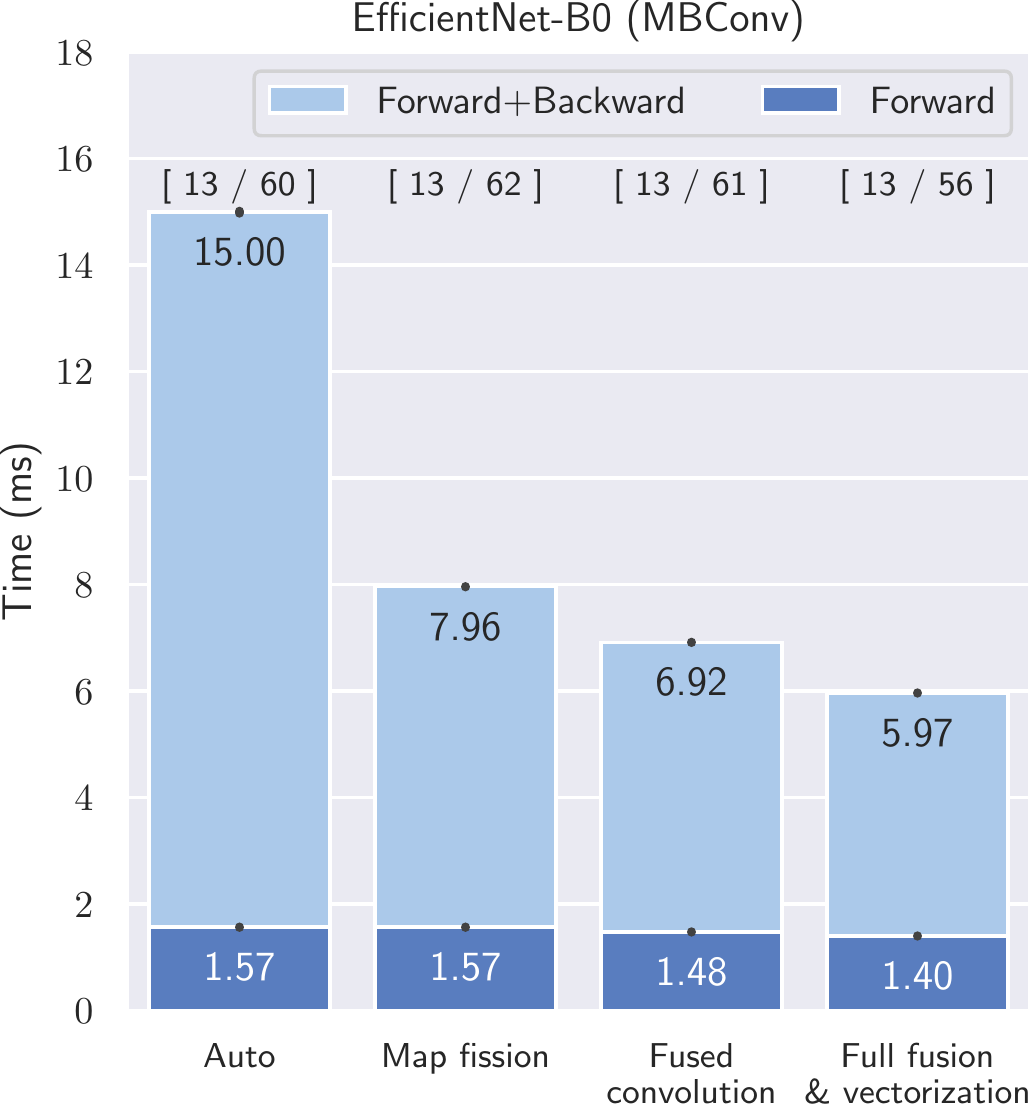}
    \vspace{-1em}
    \caption{Runtime results (left) and transformation breakdown (right) of EfficientNet-B0 guided optimization.}%
    \label{fig:eval-set1}
\end{figure}

\begin{figure}[t]
    \centering
    \includegraphics[width=0.9\linewidth]{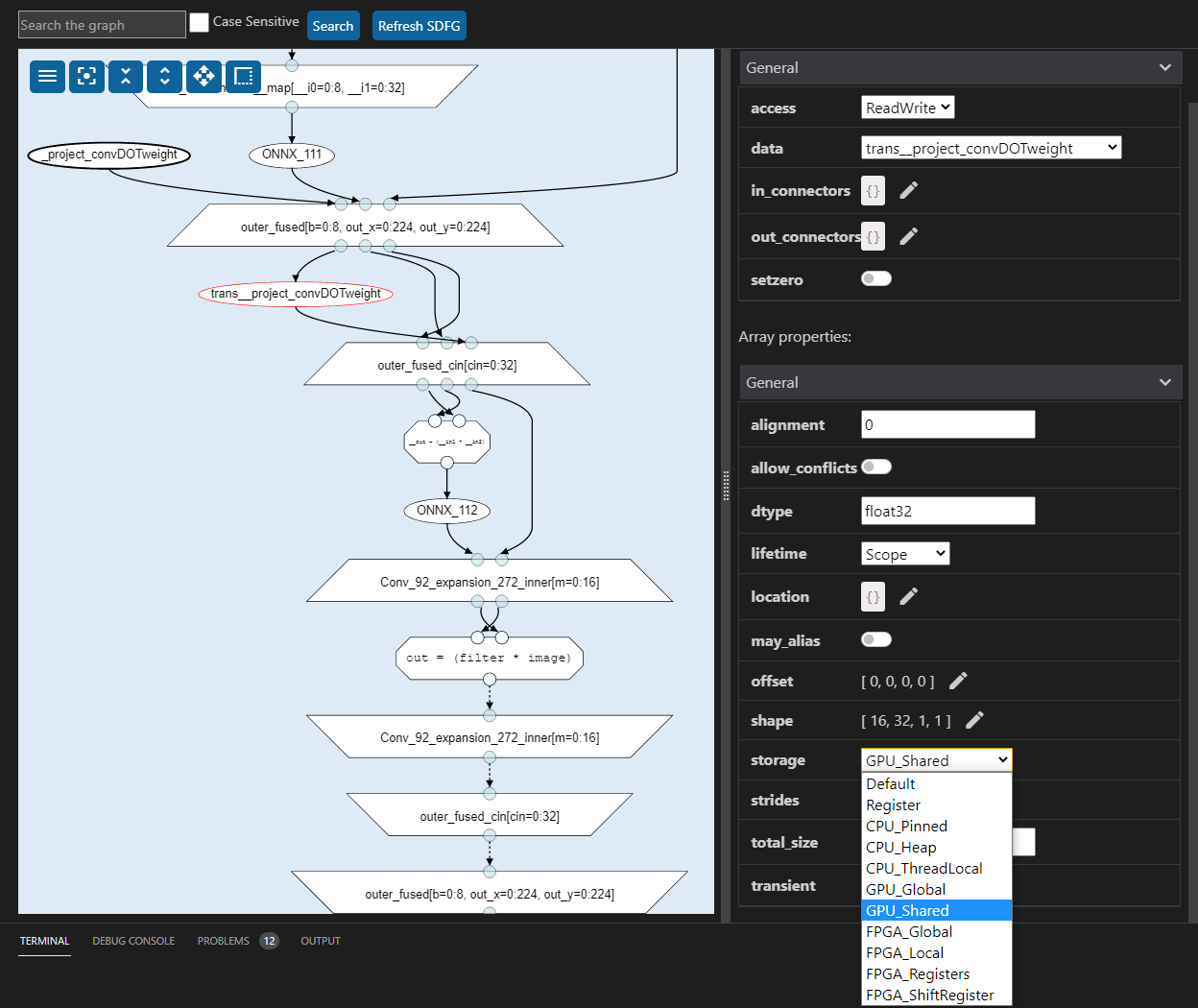}
    \vspace{-1em}
    \caption{EfficientNet optimization in Visual Studio Code.}
    \label{fig:vscode}
\end{figure}

In this case study, we consider the EfficientNet-B0~\cite{tan2019efficientnet} convolutional neural network.
Since EfficientNets use repeated blocks, we can optimize one such block and reuse the techniques for the rest of the network. We focus on the first MBConv block, where performance results are reported in Figure~\ref{fig:eval-set1}, as well as the progressive improvements gained by various optimizations, using the automatic recipe as a base that was further improved. For this case study, we perform the guided optimizations using the DaCe Visual Studio Code plugin (see Figure \ref{fig:vscode}), and report the number of clicks performed in the UI.

\textbf{Map fission} (6 clicks)\; We use two fission transformations to split a map in the backward pass. This allows us to avoid atomic operations by swapping iteration order and accumulating into thread-local memory.

\textbf{Fused convolution} (10 clicks)\; We discover a new fusion opportunity by expanding the convolutional operators to native SDFG, rather than cuDNN. Finding this automatically is non-trivial, requiring an exchange of the loop iteration variables in two operators to enable fusion. The data-centric GUI aids in this process --- viewing data movement volumes in the graph highlights a potential bottleneck even without running the program. This fused convolution nets a $\sim$1.33$\times$ speedup over cuDNN and PyTorch's hand-optimized versions in the forward pass.%

\textbf{Full fusion and vectorization} (2 clicks)\; The map fission performed enables further fusion in the backward pass. We also tune the backpropagation to recompute intermediate values rather than loading and storing them. This is done by applying \texttt{TaskletFusion} before the AD engine runs (in 27 lines of code). We fully fuse, flatten and vectorize the maps where possible.

\begin{figure}[t]
  \centering
  \includegraphics[width=0.48\linewidth]{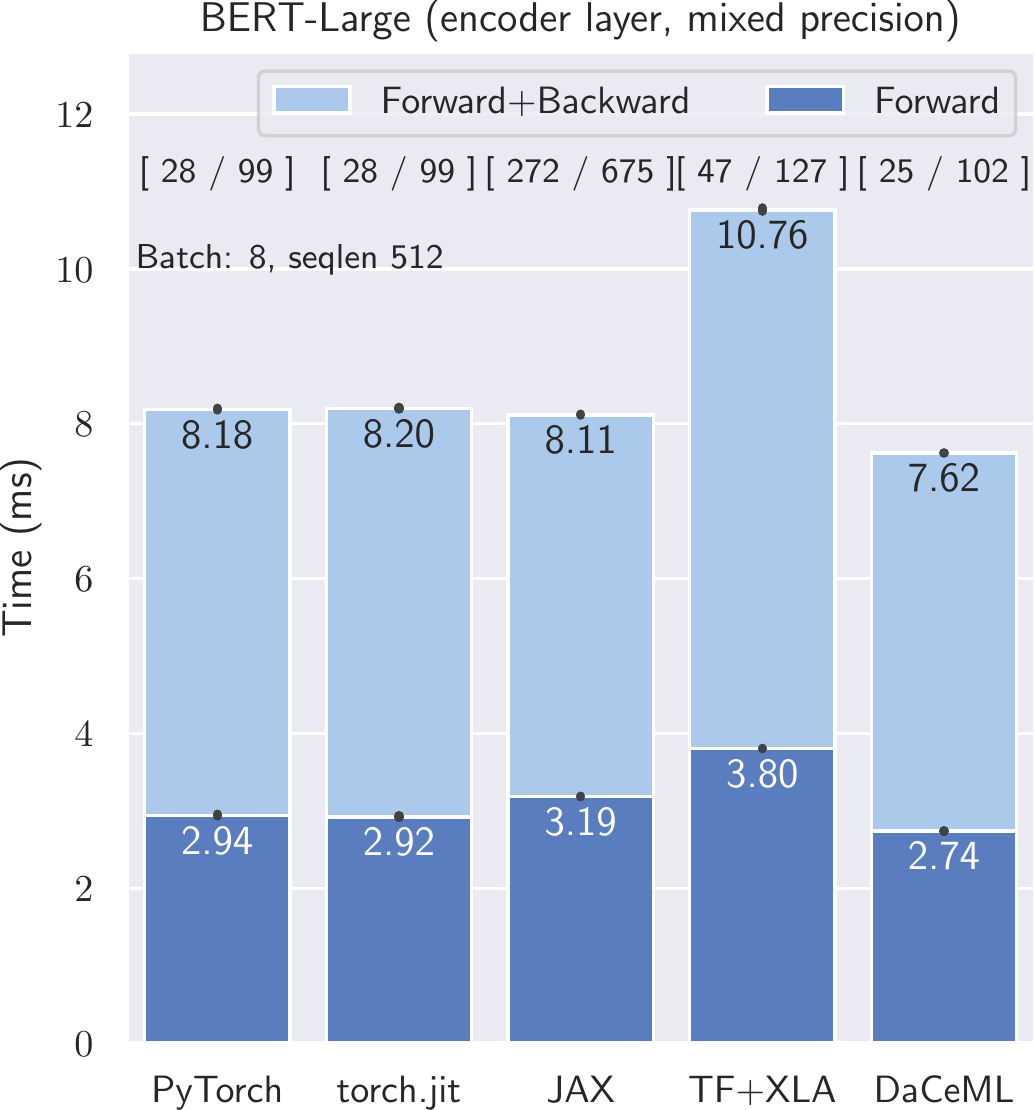}
  \hfill
  \includegraphics[width=0.48\linewidth]{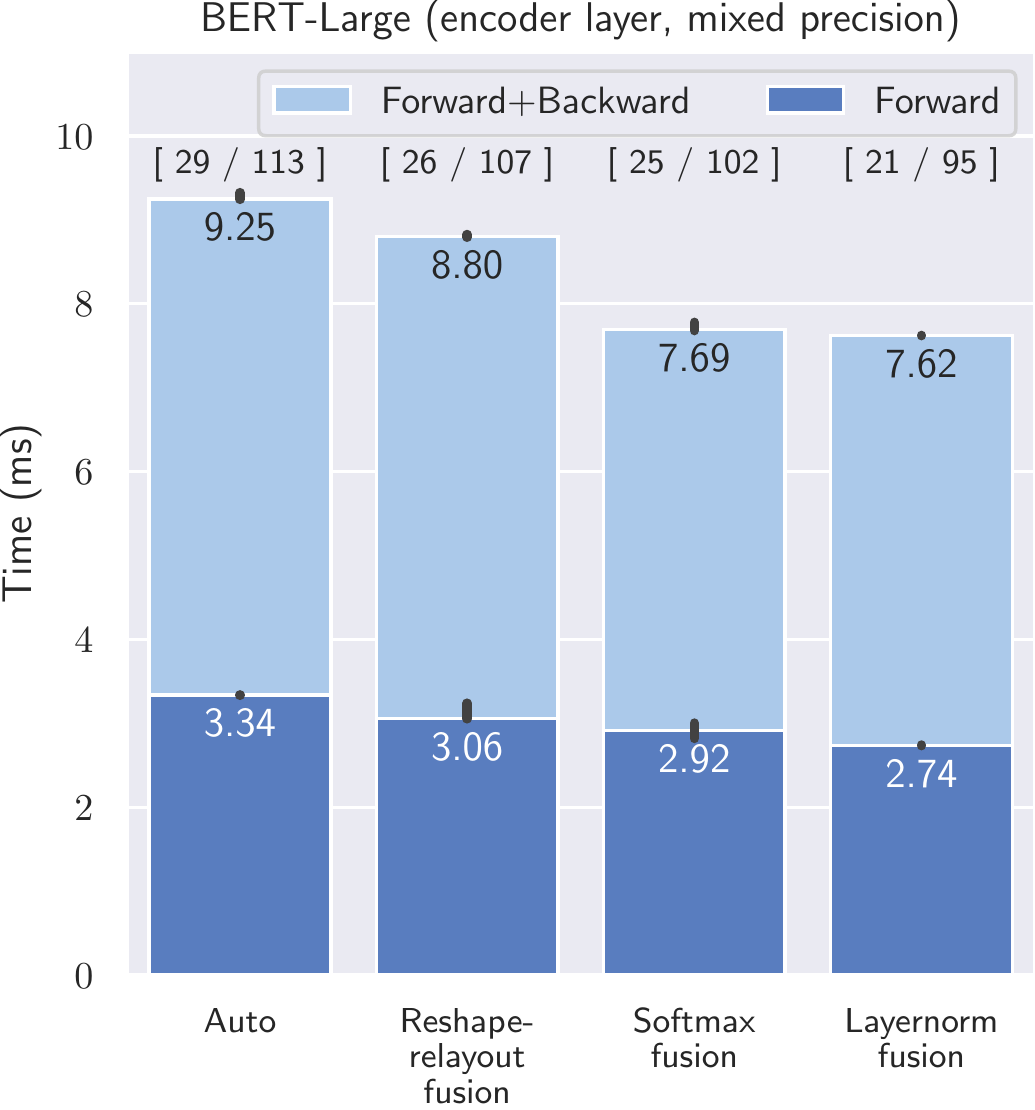}
  \vspace{-1em}
  \caption{Runtime results (left) and transformation breakdown (right) of BERT$_{\textsc{LARGE}}$ guided optimization.}
  \label{fig:eval-set2}
\end{figure}
\subsection{Guided Optimization Case Study: BERT}\label{sec:eval:bert}

Our second case study is the widely-used BERT~\cite{devlin2019bert} transformer. We optimize the BERT$_{\textsc{LARGE}}$ architecture with a batch size of 8 and sequence length 512. We run in mixed precision: here the advantages of the data-centric optimization are more pronounced as data movement becomes more important. For example, we already see in Figure~\ref{fig:eval-set2} that TF+XLA exhibits slower performance due to suboptimal data layouts that yield slower tensor contractions.

All transformations are performed using the Python API starting from the automatic recipe; we report lines of code.

\textbf{Reshape-relayout Fusion} (25 lines)\; After applying the algebraic fusion heuristic, constrained to generate BLAS operations only, we observed extraneous data relayout (transposition) calls with reshapes. We then wrote a simple transformation that detects this pattern and fuses the output write into the prior (or subsequent) computation, if elementwise. This applies six times (thrice forward, thrice backward) in the graph.

\textbf{Softmax Fusion} (39 lines)\; Instead of calling a pre-optimized library for softmax, we decide to ``break out of the library jail'' and expand it to the native SDFG, followed by applying the our generic recipe (detailed in Section~\ref{sec:eval:layernorm}). This results in fusing scaling into softmax, at a 27 $\mu$s overhead instead of 223 $\mu$s in the forward pass, and 1.1 ms gain in total.

\textbf{Layer-normalization Fusion} (47 lines)\; Here we use the lifted layer normalization scheme to nest the preceding linear layer bias into normalization. We extend and adapt the lifting transformation to include the prior/subsequent operations. For the backpropagation, since bias is reduced into 1,024 elements, we use a warp-based reduction schedule and combine it with the layer normalization weight gradient kernel, which has the same iteration space.

In both case studies, the resulting code outperforms \textit{all} compiler infrastructures. This demonstrates the strength of guided data-centric optimization --- inspecting complex DNN models from a bird's eye view for data movement bottlenecks, and mitigating them via transformations.

\section{Conclusion}
\label{sec:discussion}
We explore the concept of data-centric optimization for deep learning with DaCeML.
The framework enables general-purpose data layout and movement transformations to be applied on arbitrary networks written in PyTorch, matching and outperforming state-of-the-art compilers. The two key insights of the data-centric view are focusing on data movement minimization based on memory access patterns rather than operator types, and allowing performance engineers to further tune global and local movement. The former can perform a superset of the optimizations applied by DNN compilers; and the latter can turn massive engineering effort into a click of a button in the analysis and transformation UI.
Either automatic or human-in-the-loop, DaCeML helps practitioners speed up training without sacrificing productivity. %

\section*{Acknowledgements}
This project received funding from the European Research Council \includegraphics[height=1em]{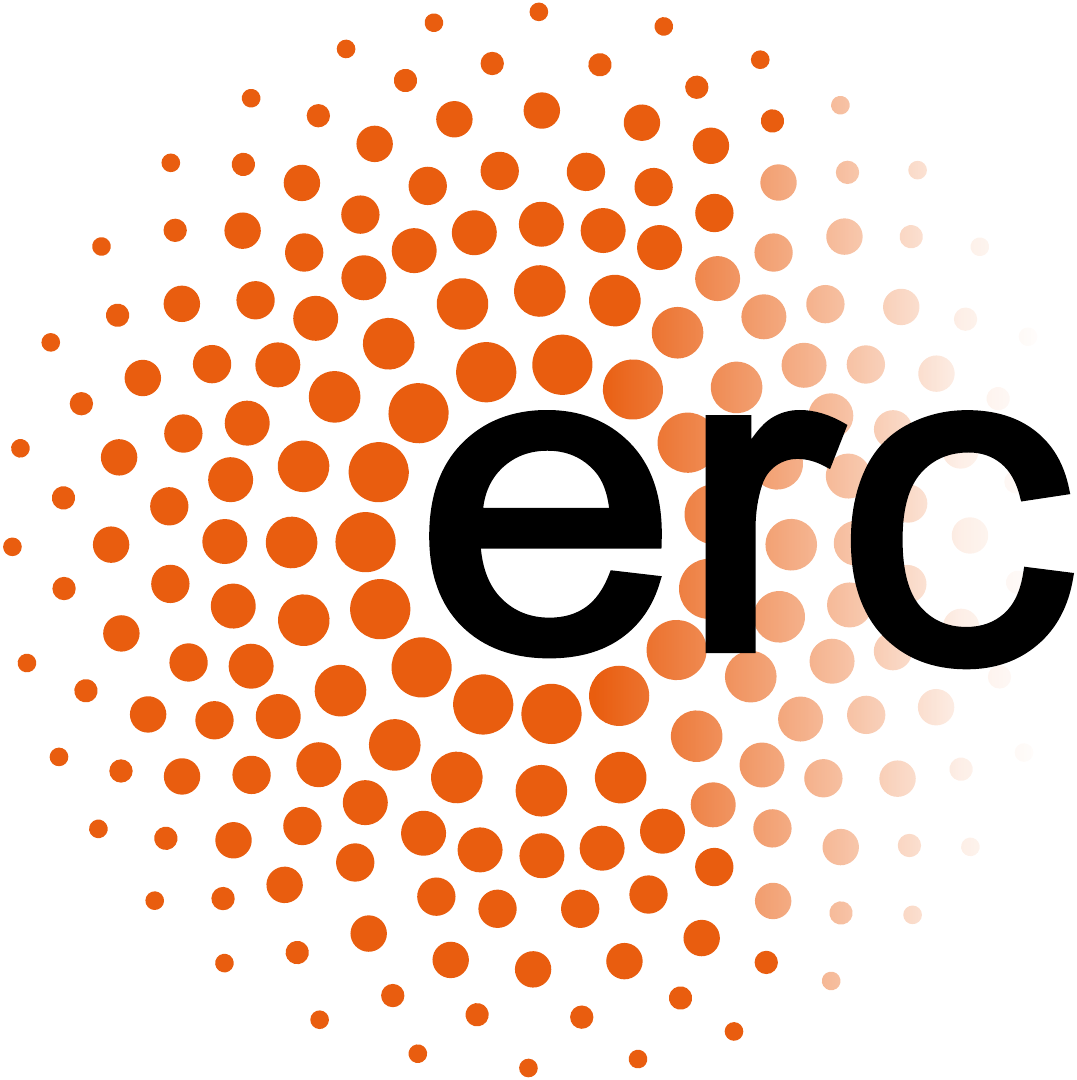} under the European Union’s Horizon
2020 programme (Project PSAP, No. 101002047); and receives EuroHPC-JU funding under grants EUPILOT, No. 101034126; MAELSTROM, No. 955513; and DEEP-SEA, No. 955606, with support from the Horizon 2020 programme. T.B.N. is supported by the Swiss National Science Foundation (Ambizione Project \#185778). N.D. is supported by the ETH Postdoctoral Fellowship. The authors wish to acknowledge the support from the PASC program (Platform for Advanced Scientific Computing), as well as the Swiss National Supercomputing Center (CSCS) for providing computing infrastructure.
\bibliographystyle{ACM-Reference-Format}
\bibliography{refs}

\end{document}